\pdfoutput=1

\documentclass{article}

\usepackage{times}
\usepackage{graphicx} 
\usepackage{subfigure} 
\usepackage{amsmath}
\usepackage{amssymb}
\usepackage{array}
\graphicspath{ {/home/mdharmo/Documents/ZPaper/} }
\usepackage{natbib}
\usepackage{tabularx,ragged2e,booktabs,caption}
\usepackage{algorithm}
\usepackage{algorithmic}

\usepackage{hyperref}

\usepackage[accepted]{icml2017} 

\icmltitlerunning{Activation Ensembles for Deep Neural Networks}

\begin{document} 

\twocolumn[
\icmltitle{Activation Ensembles for Deep Neural Networks}



\icmlsetsymbol{equal}{*}

\begin{icmlauthorlist}
\icmlauthor{Mark Harmon}{to}
\icmlauthor{Diego Klabjan}{to}
\end{icmlauthorlist}

\icmlaffiliation{to}{Northwestern University, Evanston, IL}

\icmlcorrespondingauthor{Mark Harmon}{MarkHarmon2012@u.northwestern.edu}

\icmlkeywords{deep learning, activation functions, ensemble, convex}

\vskip 0.3in
]




\begin{abstract} 
Many activation functions have been proposed in the past, but selecting an adequate one requires trial and error.  We propose a new methodology of designing activation functions within a neural network at each layer.  We call this technique an ``activation ensemble" because it allows the use of multiple activation functions at each layer.  This is done by introducing additional variables, $\alpha$, at each activation layer of a network to allow for multiple activation functions to be active at each neuron.  By design, activations with larger $\alpha$ values at a neuron is equivalent to having the largest magnitude.  Hence, those higher magnitude activations are ``chosen" by the network.  We implement the activation ensembles on a variety of datasets using an array of Feed Forward and Convolutional Neural Networks.  By using the activation ensemble, we achieve superior results compared to traditional techniques.  In addition, because of the flexibility of this methodology, we more deeply explore activation functions and the features that they capture.

\end{abstract} 

\section{Introduction}

Most of the recent advancements in the mathematics of neural networks come from four areas: network architecture \cite{jaderberg2015spatial} \cite{gulcehre2016noisy}, optimization method (AdaDelta \cite{zeiler2012adadelta} and Batch Normalization \cite{ioffe2015batch} ), activations functions, and objective functions (such as Mollifying Networks \cite{gulcehre2016mollifying} ).  Highway Networks \cite{srivastava2015training} and Residual Networks \cite{he2016deep} both use the approach of adding data from a previous layer to one further ahead for more effective learning.  On the other hand, others use Memory Networks \cite{weston2014memory} to more effectively remember past data and even answer questions about short paragraphs.  These new architectures move the field of neural networks and machine learning forward, but one of the main driving forces that brought neural networks back into popularity is the rectifier linear unit (ReLU)\cite{glorot2011deep} \cite{nair2010rectified}.  Because of trivial calculations in both forward and backward steps and its ability to more effectively help a network learn, ReLU's revolutionized the way neural networks are studied.

One technique that universally increases accuracy is ensembling multiple predictive models.  There are books and articles that explain the advantages of using multiple models rather than a single large model \cite{zhang2012ensemble}.  When neural networks garnered more popularity in the 90's and early 2000's, researchers used the same technique of ensembles \cite{zhou2002ensembling}.  Additionally, other techniques may identify as an ensemble.  For example, the nature of Dropout \cite{srivastava2014dropout} trains many different networks together by dropping nodes from the entirety of a network. 

We focus on activation functions rather than expanding the general architectures of networks.  Our work can be seen as a layer or neuron ensemble that can be applied to any variety of deep neural network via activation ensembles. We do not focus on creating yet another unique activation function, but rather ensemble a number of proven activation functions in a compelling way.  The end result is a novel activation function that is a combination of existing functions.  Each activation function, from ReLU to hyperbolic tangent contain advantages in learning.  We propose to use the best parts of each in a dynamic way decided by variables configuring contributions of each activation function.  These variables put weights on each activation function under consideration and are optimized via backpropagation.

As data passes through a deep neural network, each layer transforms the data to better interpret and gather features.  Therefore, the best possible function at the top of a network may not be optimal in the middle or bottom of a network.  The advantage of our architecture is that rather than choosing activations at specified layers or over an entire network, one can give the network the option to choose the best possible activation function of each neuron at each layer.  

Creating an ensemble of activation functions presents some challenges in how to best combine the activation functions to extract as much information from a given dataset as possible.  In the most basic model ensembles, one can average the given probabilities of multiple models.  This is only feasible because the range of values in each model are the same, which is not replicated in most activation functions.  The difficulty lies in restricting or expanding the range of these functions without losing their inherent performance. 

An activation ensemble consists of two important parts. The first is the main $\alpha$ parameter attached to each activation function for each neuron. This variable assigns a weight to each activation function considered, i.e., it designs a convex combination of activation functions.  The second are a set of ``offset" parameters, $\eta$ and $\delta$, which we use to dynamically offset normalization range for each function.  Training of these new parameters occurs during typical model training and is done through backpropagation.

Our work contains two significant contributions.  First,  we implement novel activation functions as convex combinations of known functions with interesting properties based upon current knowledge of activation functions and learning.  Second, we improve the learning of any network considered herein, including the well-established residual network in the Cifar-100 learning task.

\section{Related Work}

There is a plethora of work in building the best activation functions for neural networks.  Before ReLU activations were commonly used, deep neural networks were nearly impossible to train since the neurons would get stuck in the upper and lower areas of sigmoid and hyperbolic tangent functions. Some work has focused on improving these activation functions, such as \citet{gulcehre2016noisy}, who introduce a stochastic variable to the sigmoid and hyperbolic tangent functions.  Since sigmoid and hyperbolic tangent function both contain areas of high saturation for values in large magnitude, the stochastic variable can aid in pushing the activation functions out of high saturation areas.  In our work, rather than introducing stochasticity, we introduce several activations at each neuron, from which the network can chose a combination.  Thus, it can reap the benefits of the sigmoid and hyperbolic tangent function without being limited to these functions at each layer.   

On the other hand, \citet{clevert2015fast} create a smoother leaky ReLU function dubbed the exponential linear function in order to take advantage of including negative values.  \citet{trottier2016parametric} further generalize the exponential linear unit by Clevert.  The new exponential linear unit is called the Parametric Exponential Linear Unit (PELU):
$$
f(x) = \left\{
        \begin{array}{ll}
            \frac{a}{b} x & \quad x \ge 0 \\
            a(e^{\frac{x}{b}} -1) & \quad x \leq 0
        \end{array}
        \;a,b>0
    \right.
$$
The extra parameters, $a,b$ increase the robustness of this activation function similar to that of the leaky ReLU compared to a regular rectifier unit.  They find that this more general format dramatically increases accuracy. Our work differs from those who utilize different activation functions by being able to use as many activation functions as a user's computational capabilities allow.  Rather than choosing one for a network, we allow the network to not only choose the best function for each layer, but also for each single neuron in each of the layers.  We grant activation functions with negative values in our network, but we restrict this after a simple transformation to ensure that our functions are similar in magnitude.

\citet{li2016multi} take a different approach from typical work that focuses on activation functions.  Rather than combining inputs, they use multiple biases to find features hidden within the magnitudes of activation functions.  In this way, they can threshold various outputs to find hidden features and help filter out noise from the data.  We restrict the range of our activation functions, which helps the neural network find features that may be hidden within the magnitude of another activation function.  Thus, we are able to find hidden features via known activation functions without introducing multiple biases. 

\citet{scardapane2016learning} create an activation function during the training phase of the model.  However, they use a cubic spline interpolation rather than using the basis of the rectifier unit.  Their work differs from ours in that we use the many different available activation functions rather than creating an entirely new function via interpolation.  It is important to note that we restrict our activation function to a particular set and then allow the network to chose the best one or some combination of those available rather than have activation functions with open parameters (though this is possible in our architecture).

Maxout Networks, \citet{goodfellow2013maxout}, allow the network to choose the best activation function similar to our work.  However, Maxout Networks require many more weights to train.  For each activation function, there is an entirely new weight matrix while our extra activations only require a few parameters.  In addition, Maxout Networks find the maximum value for the activation function rather than combining the activation functions into a novel function as done in our work.  Thus, features in a Maxout Network are lost due to not being represented after the maximization function.

\citet{jin2016deep} approach the problem of activation functions through the lens of ReLU's.  Similar to PReLU's, which are leaky ReLU's with varying slope on the negative end, they create unique activation functions for rectifier units.  However, they combine the training of the network and learning of the activation functions similar to our work and that of \citet{scardapane2016learning}.  Their work restricts the combination to a set of linear functions with open parameters, while our does not have a restriction on the type of functions.  Since the work of \citet{jin2016deep} has this restriction, they cannot combine functions of various magnitude like the work we present here.

\citet{chen2016combinatorially} uses multiple activation functions in a neural network for each neuron in the field of stochastic control.  Similar to our work, he combines functions such as the ReLU, hyperbolic tangent, and sigmoid.  To train the network, Chen uses Neuroevolution of Augmenting Topologies (NEAT) to train his neural network for control purposes.  However, he simply adds together the activation functions without capturing magnitude and does not allow the network to choose an optimal set of activations for each neuron.

Work by \citet{agostinelli2014learning}, which is closest to our approach, explores the construction of activation functions during the training phase of a neural network.  The general framework of their activation function is the following:

\begin{equation*}
\begin{aligned}
& & h_i (x) = \max (0,x) + \sum_{s=1}^{S} a_{i}^{s} \max (0,-s + b_i^s) \\  
\end{aligned}
\end{equation*}

where $i$ denotes the neuron, $S$ is a hyperparameter, and $a$ and $b$ are trained variables chosen by the user.  Thus, rather than combining various activations functions together via an optimization process, \citet{agostinelli2014learning} take the baseline foundation of the ReLU function and add additional training variables to the function through $a$ and $b$.  The difference in our work is that we want to capture the features extracted by the various activation functions actively used to find the best combination available.    

\section{Activation Ensembles}

Ensemble Layers were created with the idea of allowing a network to choose its own activation for each neuron and for each layer of the network.  Overall, the network takes the output of a previous layer, for example from a convolutional step, applies its various activations, normalizes these activations, and places weights on each activation function.  We first go through each step of the process of making such a layer.

The first naive approach is to simply take the input, use a variety of activation functions, and add these activation functions together. We denote the set of activation functions we use to train a network $f^j \in F$.  Using this method, a network may reap the benefits of having more than one activation function, which may extract different features from the input.  However, simply adding poses a problem for most functions.  Many functions, like the sigmoid and hyperbolic tangent possess different values, but they can be easily scaled to have the same range.  However, other functions, such as the rectifier family and the inverse absolute value function cannot be easily adjusted due to their unbounded ranges.  If we simply add together the functions, the activation functions with the largest absolute value will dominate the network leaving the other functions with minimal input.

To solve this issue, we need to normalize the activation functions with respect to one another in order to have relatively equal contribution to learning.  One option would be to use mean and standard deviation normalization; however, this would not equalize contribution.  Therefore, we scale the functions to [0,1].  While building our method, we additionally performed tests using the range of [-1,1] for each activation function.  We found that the performance was either close to that of [0,1] or slightly worse.  In addition, allowing negative values causes additional issues when choosing the best activation functions with the $\alpha$ parameter we introduce later.

Simply adding activations together and forcing them to have equal contribution does not solve our second goal of finding the best possible activation functions for particular problems, networks, and layers.  Therefore, we apply an additional weight value, $\alpha$, to each activation for each neuron.  Therefore, for the output of each neuron $i$ and $m$ being the number of activation functions, we have the following activation function for each neuron:

\begin{equation*}
\begin{aligned}
& h_{i}^j (z) = \frac{f^j (z) - \min_k (f^j(z_{ki}))}{\max_k f^j(z_{ki}) - \min_k f^j(z_{ki}) + \epsilon} \\
& y_i (z) = \sum_{j=1}^m \alpha_{i}^j h_{i}^j (z)\\
\end{aligned}
\end{equation*}

Here, $\epsilon$ is a small number, $k$ goes through all training samples (in practice $k$ varies over the samples in a minibatch), and $z_{ki}$ is the input to neuron $i$ with training example $k$. We consider $\alpha_{i}^j$ as part of our network optimization. In order to again avoid one activation growing much larger than the others we must also limit the magnitude of $\alpha$.  Since the activation functions are contained to [0,1], the most obvious choice is to limit the values of $\alpha$ to lie within the same range.  Thus, the network also has the ability to choose a particular activation function $f^j\in F$ if the performance of one far outweighs the others.  For example, during our experiments, there were many neurons that we found heavily favored the ReLU function after training signified by the $\alpha^j$ for the ReLU function being magnitudes larger than the others.

In order to force the network to choose amongst the presented activation functions, we further require than all the weights for each neuron add to one.  This then gives us another optimization problem to solve when updating the weights. In what follows, the values $\hat{\alpha}^j$ are the proposed weight values for the activation function $j$ gradient update and $\alpha^j$ are the new values we find after solving the projection problem.  We omit the neuron index $i$ for simplicity, i.e. this problem must be solved for each neuron.

\begin{equation*}
\begin{aligned}
& \underset{\alpha}{\text{minimize}}
& & \sum_{j=1}^{m} \tfrac{1}{2} (\hat{\alpha^j} - \alpha^j)^2 \\
& \text{subject to}
& & \sum_{j=1}^{m} \alpha^j  = 1 \\
&&& \alpha^j \geq 0,  \; j=1,2, \ldots m.  
\end{aligned}
\end{equation*}

This optimization problem is convex, and is readily solvable via KKT conditions.  It yields the following solution:

\begin{equation*}
\begin{aligned}
&  \alpha^j=\begin{cases}
    \hat{\alpha}^j + \lambda, & \text{if $\lambda > \alpha^j$},\\
    0, & \text{otherwise}.
  \end{cases}\\
& \sum_{j=1}^{m} \max(0,\hat{\alpha}^j + \lambda)=1  
\end{aligned}
\end{equation*}

This problem is very similar to the water-filling problem, and has an $O(m)$ solution.  Algorithm 1 exhibits the procedure for solving this problem.  The projection sub-problem must be solved for each neuron for each layer that our multiple activations are applied.

\begin{algorithm}[h]
   \caption{Projection Sub-Problem}
   \label{alg:1}
\begin{algorithmic}
   \STATE {\bfseries Input:} Vector $\hat{\alpha}$, 
   \STATE Initialize Vector $G = 1$ and Scalar $D=m$  
   \FOR{$k=1$ {\bfseries to} $m$}
   \STATE $\lambda=\frac{1-\sum_{j=1}^{m} \hat{\alpha}^j}{D}$
   \STATE $\alpha = \hat{\alpha} + \lambda$
   \STATE $\alpha = \alpha\circ G$ (Hadamard Product)
   \IF{any $\alpha^k < 0$} 
   \STATE $D = D-1$
   \STATE $G_{k} = 0$
   \ENDIF
   \ENDFOR
\end{algorithmic}
\end{algorithm}
 
We borrow two elements in our approach from Batch Normalization \cite{ioffe2015batch}.  We record running minimum and maximum values while training over each minibatch.  Thus we transform the data using only a small portion (dictated by batch-size during training).  The other element that we introduce is the two parameters, $\eta$ and $\delta$, which allow for the possibility of the network choosing to leave the activation at its original state.  Below is the final equation that we use for the activation function at each neuron.

\begin{equation*}
\begin{aligned}
& y_i = \sum_{j=1}^m \alpha^j(\eta^j h_{i}^j + \delta^j) 
\end{aligned}
\end{equation*}

Therefore, the final algorithm for our network involves both maintaining running maximum and minimum values and solving the projection subproblem during training.  Then in the test phase, we use the approximate minimum and maximum values we obtain during the training phase.  The weights learned during the training phase, namely our new parameters $\alpha,\eta,$ and $\delta$ remain as is during training. In summary, rather than creating a new weight matrix for each activation function, which adds a huge amount of overhead, we allow the network to change its weights according the activation function that is optimal for each neuron.

We next provide derivatives for the new parameters.  Let $\ell$ denote the cost function for our neural network.  We backpropagate through our loss function $\ell$ with the chain rule to find the following:

\begin{equation*}
\begin{aligned}
\frac{\partial \ell}{\partial \alpha^j} &= \frac{\partial \ell}{\partial y_i} \cdot \Big(\eta^j h_{i}^j + \delta^j \Big)\\
\frac{\partial \ell}{\partial \eta^j} &= \frac{\partial \ell}{\partial y_i} \cdot \alpha^j h_{i}^j&&\\
\frac{\partial \ell}{\partial \delta^j} &= \frac{\partial \ell}{\partial y_i}\cdot \alpha^j 
\end{aligned}
\end{equation*}

As a note, activation ensembles and the algorithms above work in the same way for CNN's.  The only difference is that instead of using a set of activations for a specific neuron, we use a set of activations for a specific filter.

\subsection{Activation Sets}

To explore the strength of our ensemble method, we create three sets of activations to take advantage of the weakness of individual functions. The first set is a number of activation functions seen in networks today.  One of the functions, the exponential linear units, garners favorable results with datasets such as CIFAR-100. Others include the less popular inverse absolute value function and the sigmoid function (which is primarily relegated to recurrent networks in most literature).  One omission we must mention is the adaptive piecewise linear unit. Since our goal is to create an activation from common functions, a function that focuses on adapting via weights is not included; however, this does not mean it would not work within our model.

\begin{itemize}  
\item Sigmoid Function: $\frac{1}{1+e^{-z}}$
\item Hyperbolic Tangent: $\tanh(z)$
\item Soft Linear Rectifier: $\ln (1 + e^{-z})$
\item Linear Rectifier: $\max(0,z)$
\item Inverse Absolute Value: $\frac{z}{1+|z|}$
\item Exponential Linear Function:  $$
 \left\{
        \begin{array}{ll}
            z & \quad z \ge 0 \\
            e^{z} -1 & \quad z \leq 0
        \end{array}
    \right.
$$
\end{itemize}
The next two sets of activations are designed to take advantage of the ensemble method's ability to join similar functions to create a better function for each neuron.  Since ReLU functions are widely regarded as the best performing activations functions for most datasets and network configurations, we introduce an ensemble of varying ReLU functions.  Since rectifier neurons can "die" when the value drops below zero, our ensemble uses rectifiers with various intercepts of form $f^b (z) = \max(0,z+b)$, where $b\in \mathbb{R}$.  We settle with five values for $b$, $[-1.0, -0.5, 0, 0.5, 1.0]$, to balance around the traditional rectifier unit.

The final activation set is a reformation of the absolute value function.  It only consists of two mirrored ReLu functions of the form:
\begin{equation*}
\begin{aligned}
& f^1 (z) = \max(0,-z)\\
& f^2 (z) = \max(0,z)\\
\end{aligned}
\end{equation*}
The behavior of the graphed function is very similar to that of an absolute value; however, our function contains individual slopes that vary by the value of $\alpha_1$ and $\alpha_2$.  We create this function to capture elements of the data that may not necessarily react positively with respect to a rectifier unit, but still carry important information for prediction.  

\section{Experiments}

For our experiments, we use the datasets MNIST, ISOLET, CIFAR-100, and STL-10.  For MNIST and ISOLET, we use small designed feed forward and convolutional neural networks.  For CIFAR-100, we apply a residual neural network.  Last, we design an auto-encoder strategy for STL-10.  In addition, we use Theano and Titan X GPU's for all experiments.  We found that the optimization function AdaDelta performs the best for all of the datasets.  For each network, we set the learning rate for AdaDelta to 1.0, which is the suggested value by the authors for most cases.  In each network that was designed in-house, we implemented batch normalization before each ensemble layer. We describe the architecture used for each dataset under their respective sections. The nomenclature we use for network descriptions is $(16)3c-2p-10f$ where $(16)3c$ describes a convolutional layer with 16 filters of size 3x3, $2p$ for a max-pooling layer with filter size 2x2, and $10f$ for 10 neurons of a feed foward layer.

The weights at the ensemble layer were initialized at $\frac{1}{m}$ where $m$ is the number of activation functions at a neuron. Furthermore, we set $\eta = 0$ and $\delta=1$ and initialized the traditional neural network weights with the Glorot method.  In addition, we initialize the residual network using the He Normal method.  We train on our three activation sets as well as the same networks with rectifier units for the original networks since they are the standard in most cases.  Our stopping criterion is based upon the validation error for each network except for the residual network, in which the suggested number of epochs is 82.  Also, we apply AdaDelta for each optimization step for all new variables as well, $\alpha, \eta,$ and $\delta$ for each minibatch.  In Table 1 below, we summarize the test accuracy (reconstruction loss for STL-10) of our datasets and various models.  Each number is an average over five runs with different random seeds.  Note that the largest improvement is found for the ISOLET dataset.

\begin{minipage}{\linewidth}
\centering
\captionof{table}{Final Model Results for Models with and without Activation Ensembles} \label{tab:title} 
\begin{tabular}{llr}
\multicolumn{2}{c}{} \\
\cline{1-3}
DataSet    & Model & Original/Ensemble \\
\hline
MNIST      & FFN    & 97.73\% / 98.37\%      \\
MNIST     &CNN      & 99.34\% / 99.40\%            \\
ISOLET      & FFN     & 95.16\% / 96.28\%        \\
CIFAR-100   & Residual     & 73.64\% / 74.20\%      \\
STL-10       & CAE     & 0.03524 / 0.03502      \\
\lasthline
\end{tabular}
\end{minipage}

\subsection{Comparing Activation Functions}

We first explore the $\alpha$ parameter values of our activation functions.  We primarily concentrate on the first set of activations (Sigmoid, Tanh, ReLU, Soft ReLu, ExpLin, InvAbs).  Since ReLU is the most common activation function in literature, we expect it be chosen the most by our networks.  As seen Figures 1,3,4, and 5, we find this to be true.  However, neurons that are deeper may not choose any particular activation.  In fact, at some neurons in the bottom layers, the parameters for choosing a function are nearly equal.    

Figure 1 illustrates the differences between layers of our activation ensembles for the in-house Feed-Forward Network model applied to the MNIST dataset.  Each data point is taken at the end of an epoch during training.  The neurons in the images were chosen randomly.

\begin{figure}[h]
  \centering
  \begin{minipage}[b]{0.22\textwidth}
    \includegraphics[width=\textwidth]{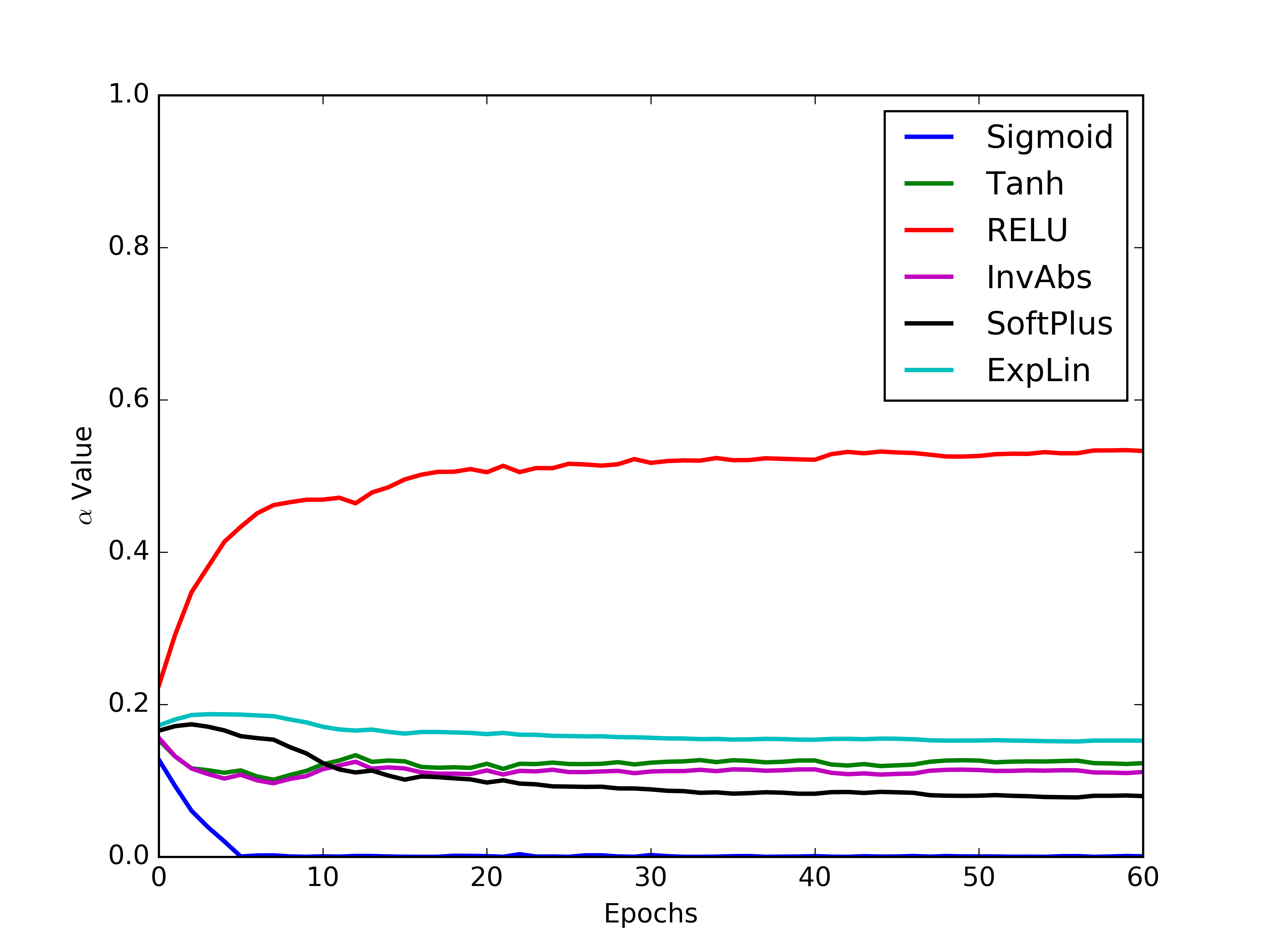}
  \end{minipage}
  \begin{minipage}[b]{0.22\textwidth}
    \includegraphics[width=\textwidth]{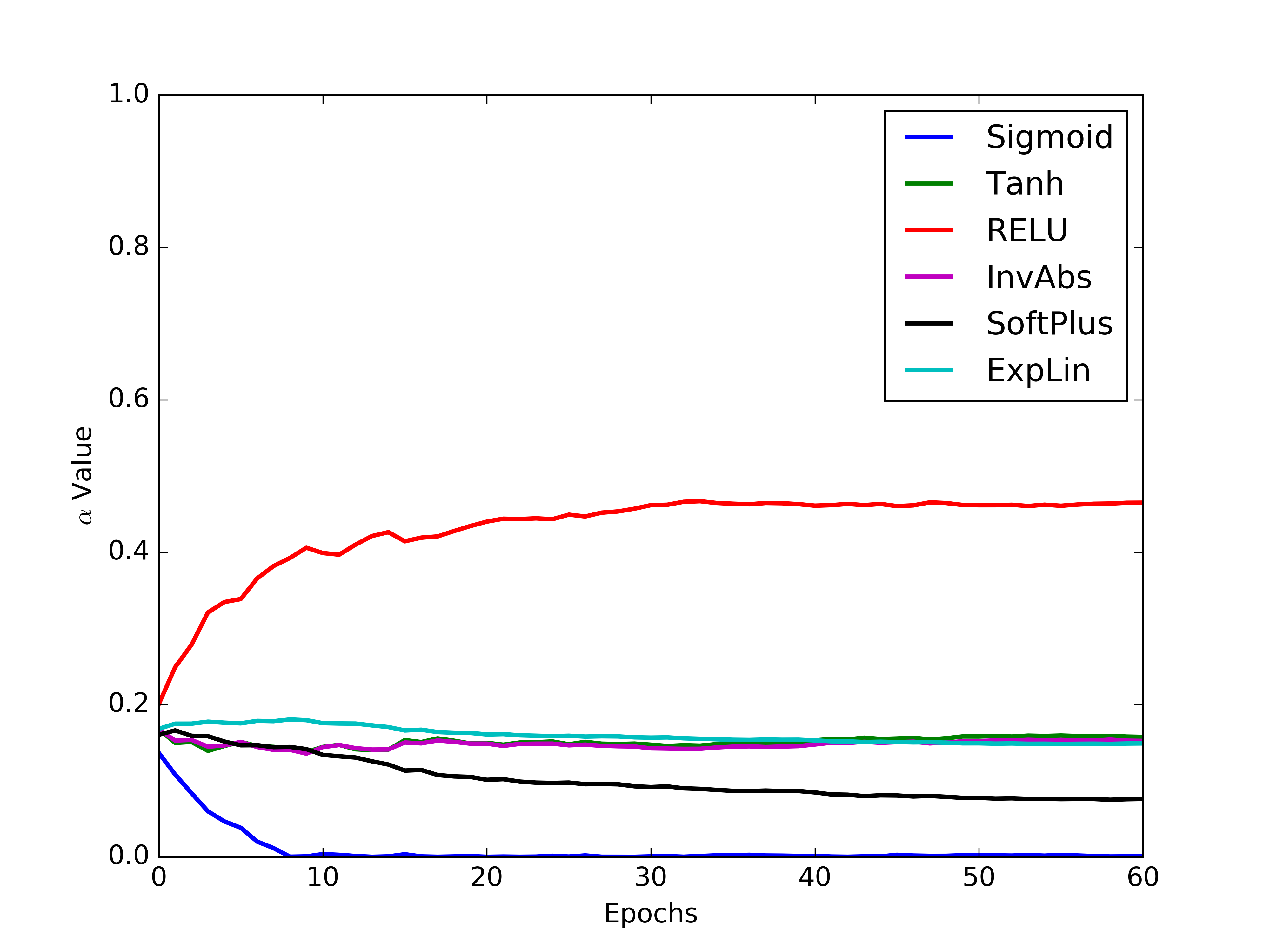}
  \end{minipage}
  \begin{minipage}[b]{0.22\textwidth}
    \includegraphics[width=\textwidth]{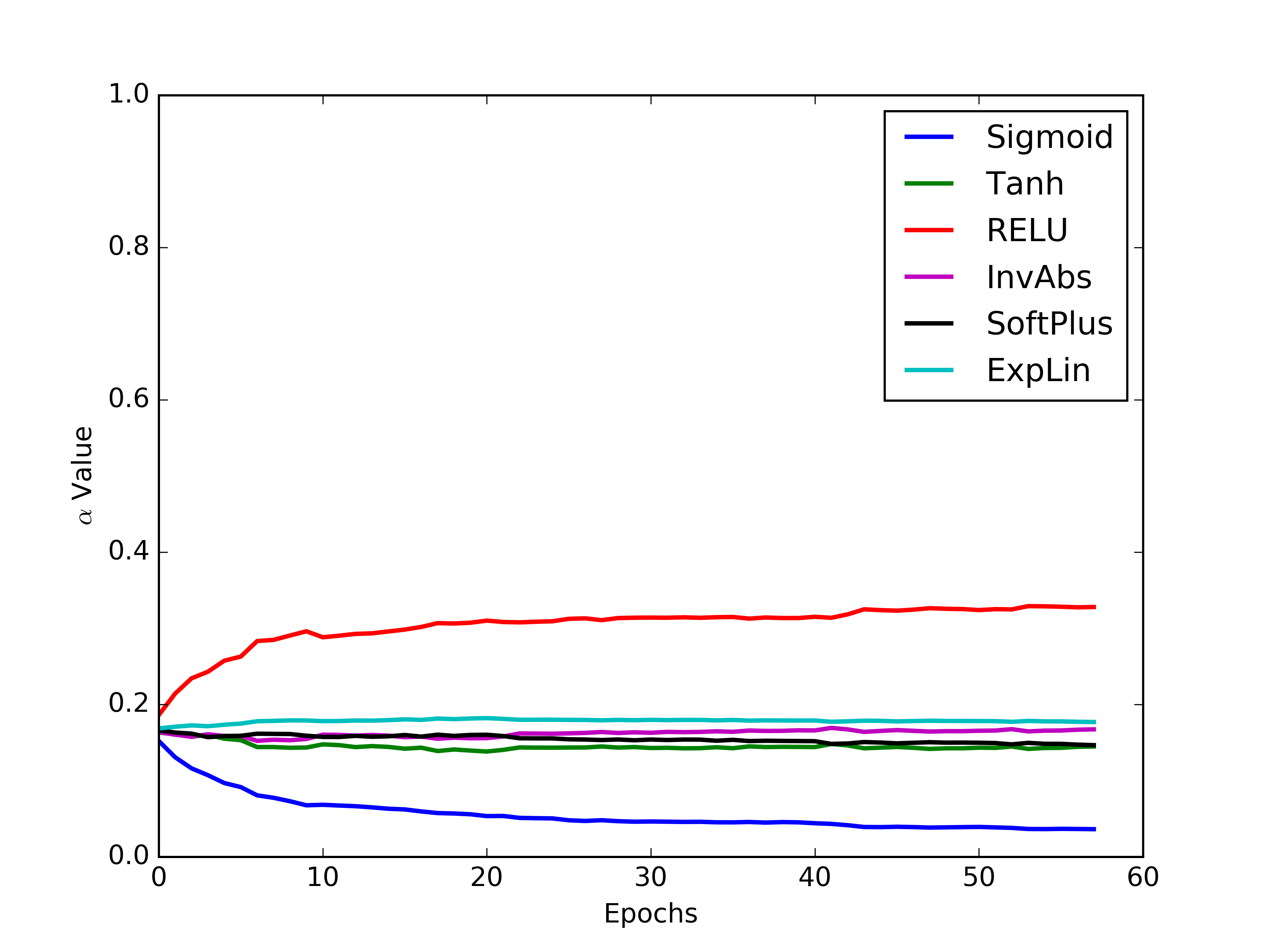}
  \end{minipage}
  \begin{minipage}[b]{0.22\textwidth}
    \includegraphics[width=\textwidth]{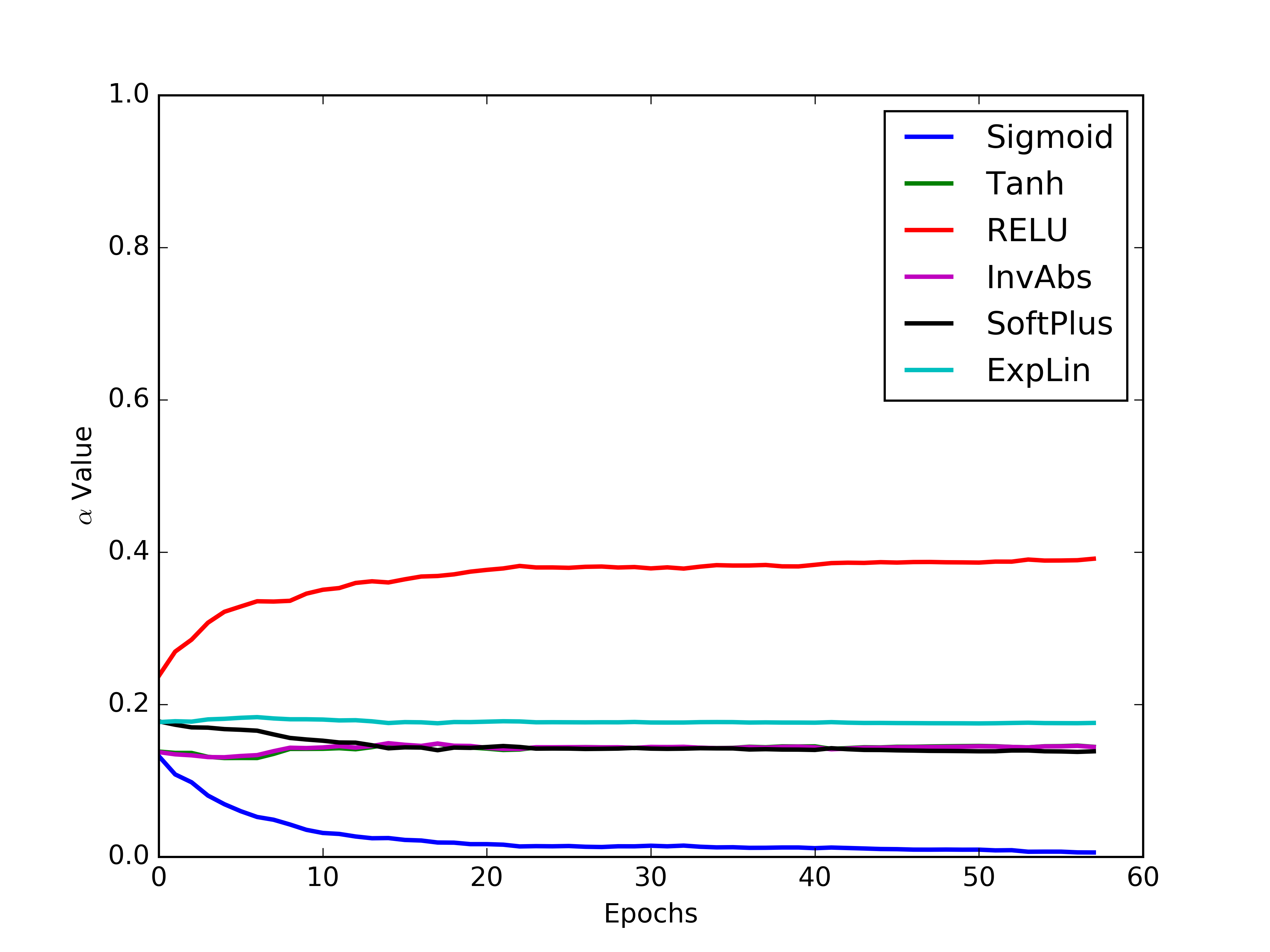}
  \end{minipage}  
  \caption{Top two images are neurons from the first layer while the bottom two are from the second layer}
\end{figure}

The top two images if Figure 1, which come from the top layer of the network, show that the ReLU becomes the leading function very quickly.  The other functions, with the exception of the softplus and exponential linear functions, rapidly approach zero.  The bottom two images, which are neurons from the next layer in the network, still choose the rectifier unit, but not as resolutely as in the previous layer.  It is interesting that the inverse absolute value function becomes very important.  We experimented as well with momentum, and found that with a step of $0.1$, the network still chooses a leading activation function.  We find that this is a general trend for FFN's and the MNIST dataset.  However, as we discuss below, this does not hold for all models or datasets.

\begin{figure}[h]
  \centering
  \begin{minipage}[b]{0.22\textwidth}
    \includegraphics[width=\textwidth]{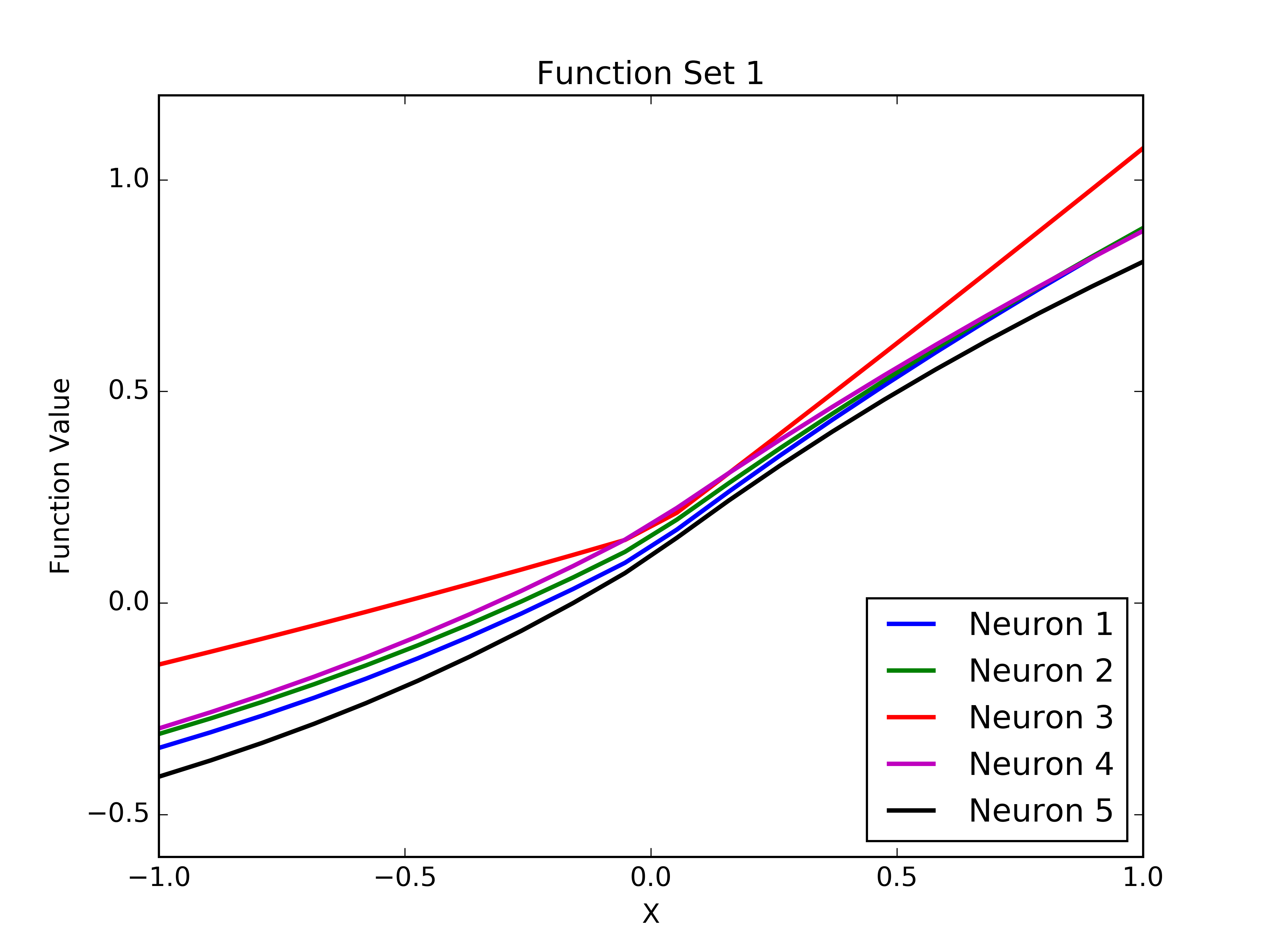}
  \end{minipage}
  \begin{minipage}[b]{0.22\textwidth}
    \includegraphics[width=\textwidth]{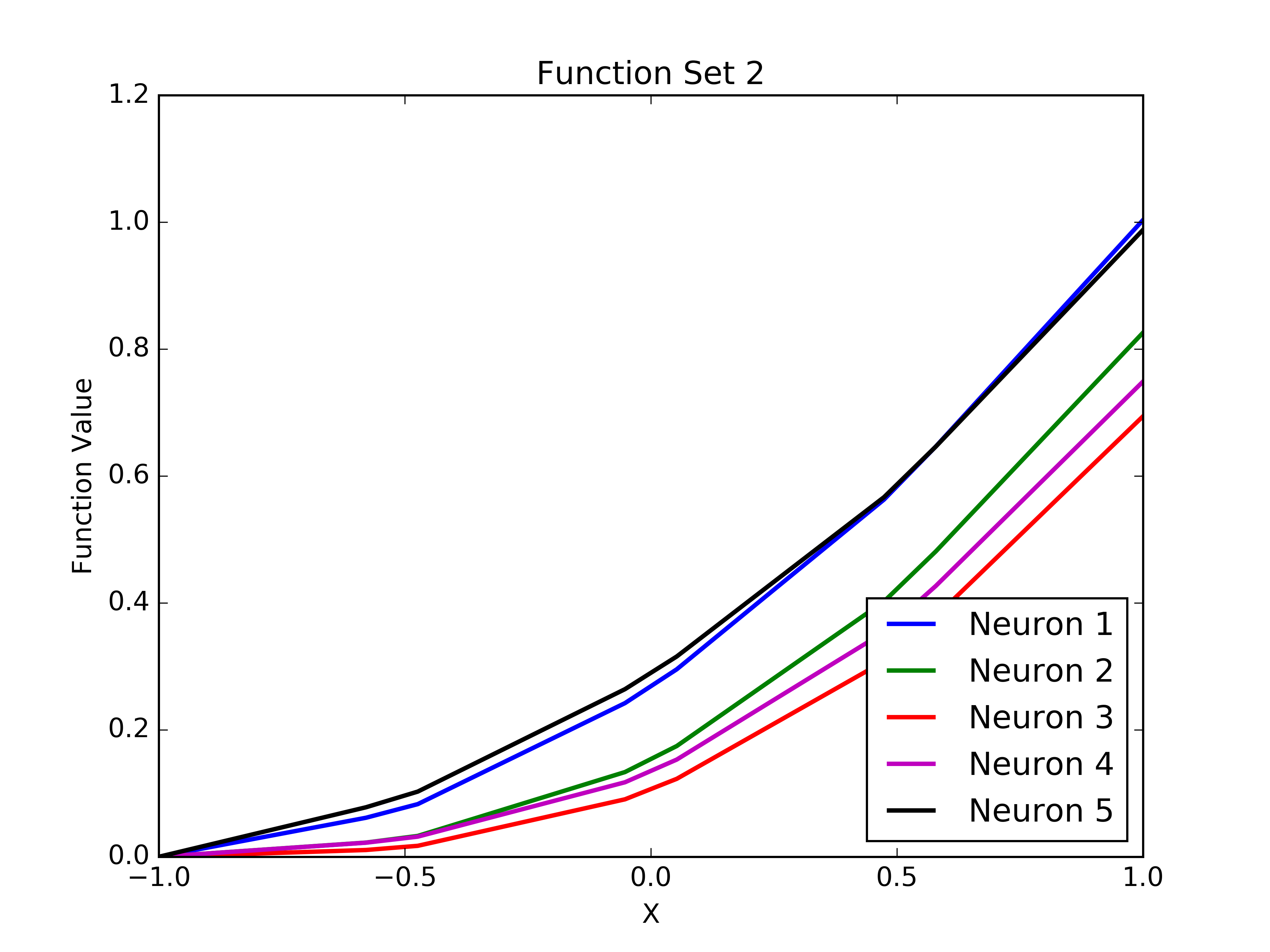}
  \end{minipage}
  \begin{minipage}[b]{0.22\textwidth}
    \includegraphics[width=\textwidth]{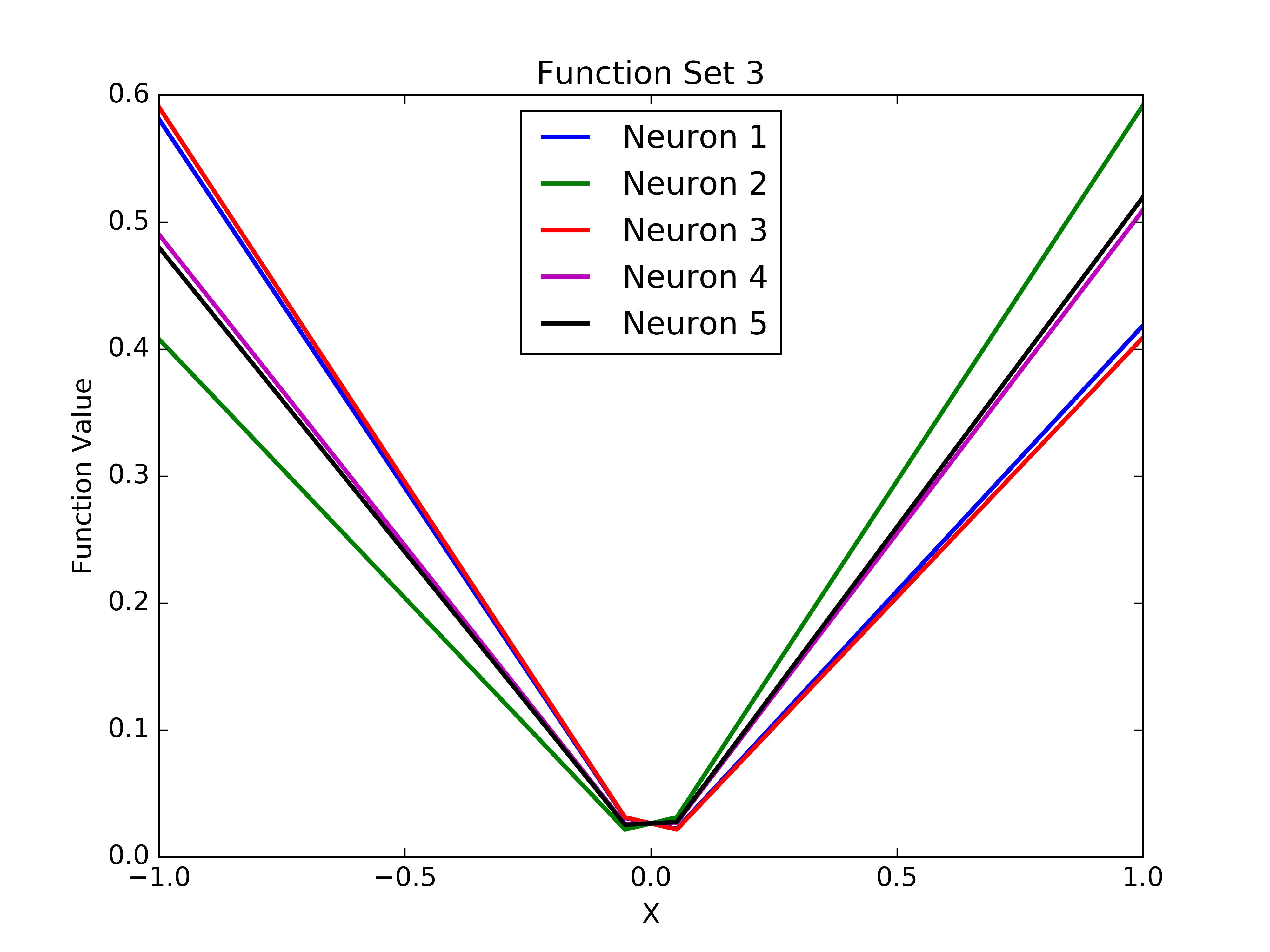}
  \end{minipage} 
  \caption{Example activation functions from each of the three sets of functions.}
\end{figure}

Next we compare the functions that the ensembles create. Figure 2 depicts the various sets of activations for our ensemble. The images are of the same five random neurons taken from the second layer of the FFN for the MNIST dataset in the first two figures while the final image is taken from the second layer of the same model trained on the ISOLET dataset.  The images are made by taking the final $\alpha$'s for each activation function, inputing a uniform vector from $[-1,1]$, and adding the functions together.  Note that we leave out the normalization technique used for activation ensembles and the values for $\eta$ and $\delta$.  The first set of activations, presented in the top left of Figure 2, behave very similarly to the exponential linear unit.  However, this function appears closer to an $x^3$ function because of its inflection point near the origin.  The second set of functions clearly form a piecewise ReLU unit with the various pieces clearly visible.  The last set appears much like an absolute value function, but differentiates itself by having varying slopes on either side of the y-axis. 

We observe that the best activation function for each model is the third set, with the exception of the residual network.

\subsection{MNIST}

We solve the MNIST dataset using two networks, a FFN and a CNN.  The FFN has the form $400f-400f-400f-10f$ while the CNN is of the form $(32)3c-2p-(32)3c-2p-400f-10f$.  MNIST is not a particularly difficult dataset to solve and is one of the few image-based problems that classical FFN's can solve with ease.  Thus, this problem is of particular interest since we may compare results between two models that can easily solve the classification problem with our activation ensembles.  One of the aspects we explore is whether or not different activations fancy certain models.

\begin{figure}[h]
  \centering
  \begin{minipage}[b]{0.22\textwidth}
    \includegraphics[width=\textwidth]{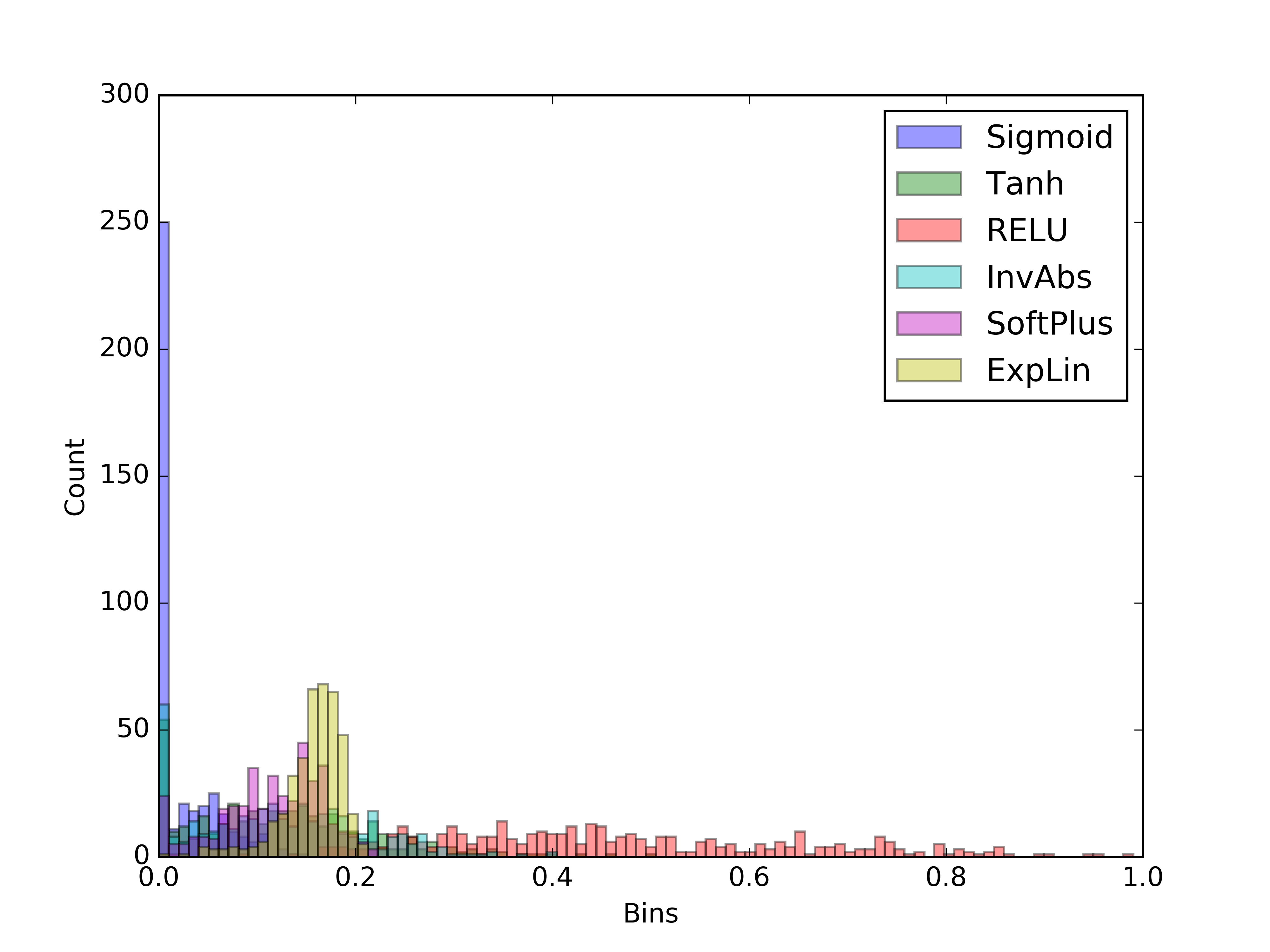}
  \end{minipage}
  \begin{minipage}[b]{0.22\textwidth}
    \includegraphics[width=\textwidth]{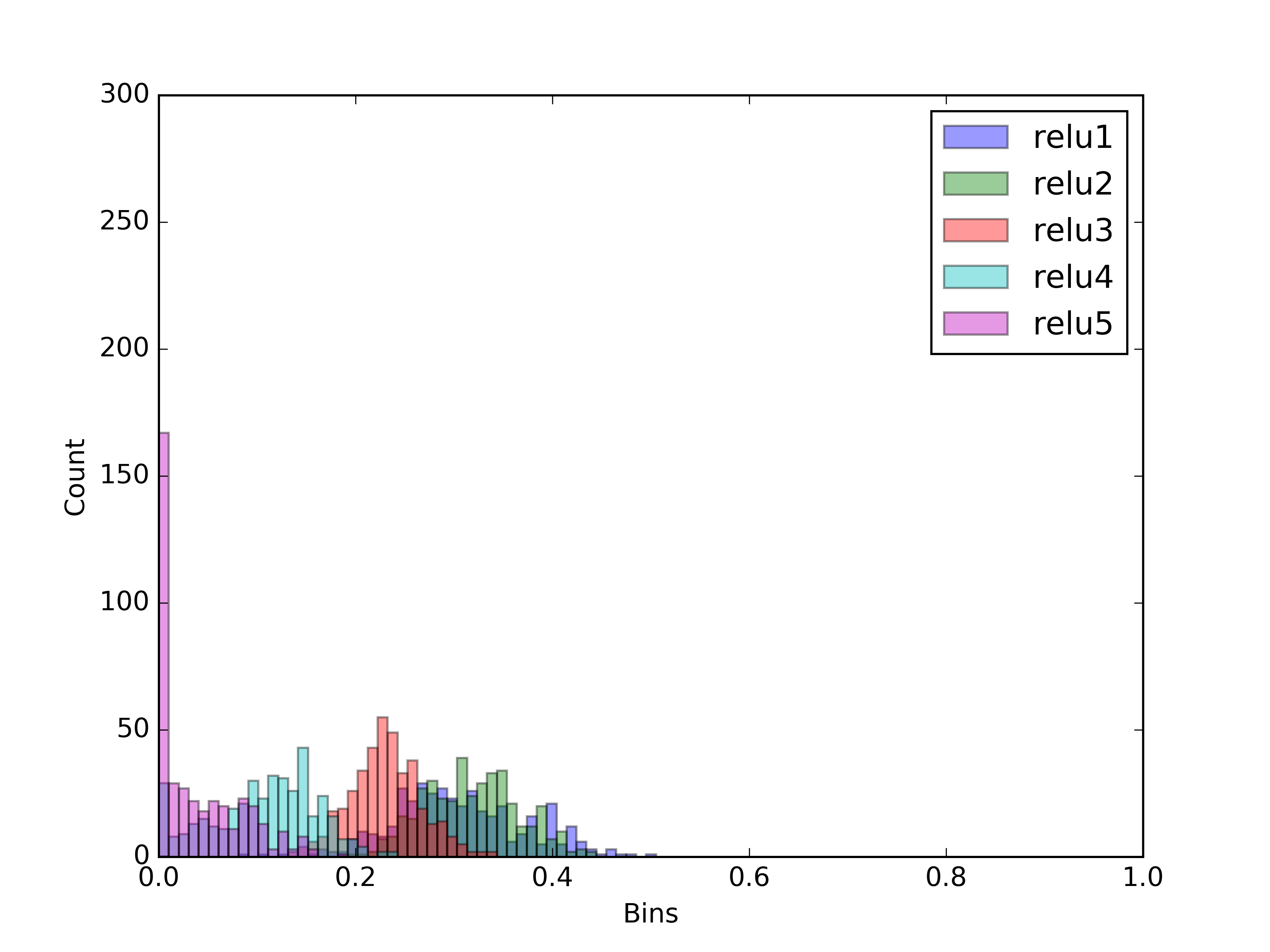}
  \end{minipage}
    \begin{minipage}[b]{0.22\textwidth}
    \includegraphics[width=\textwidth]{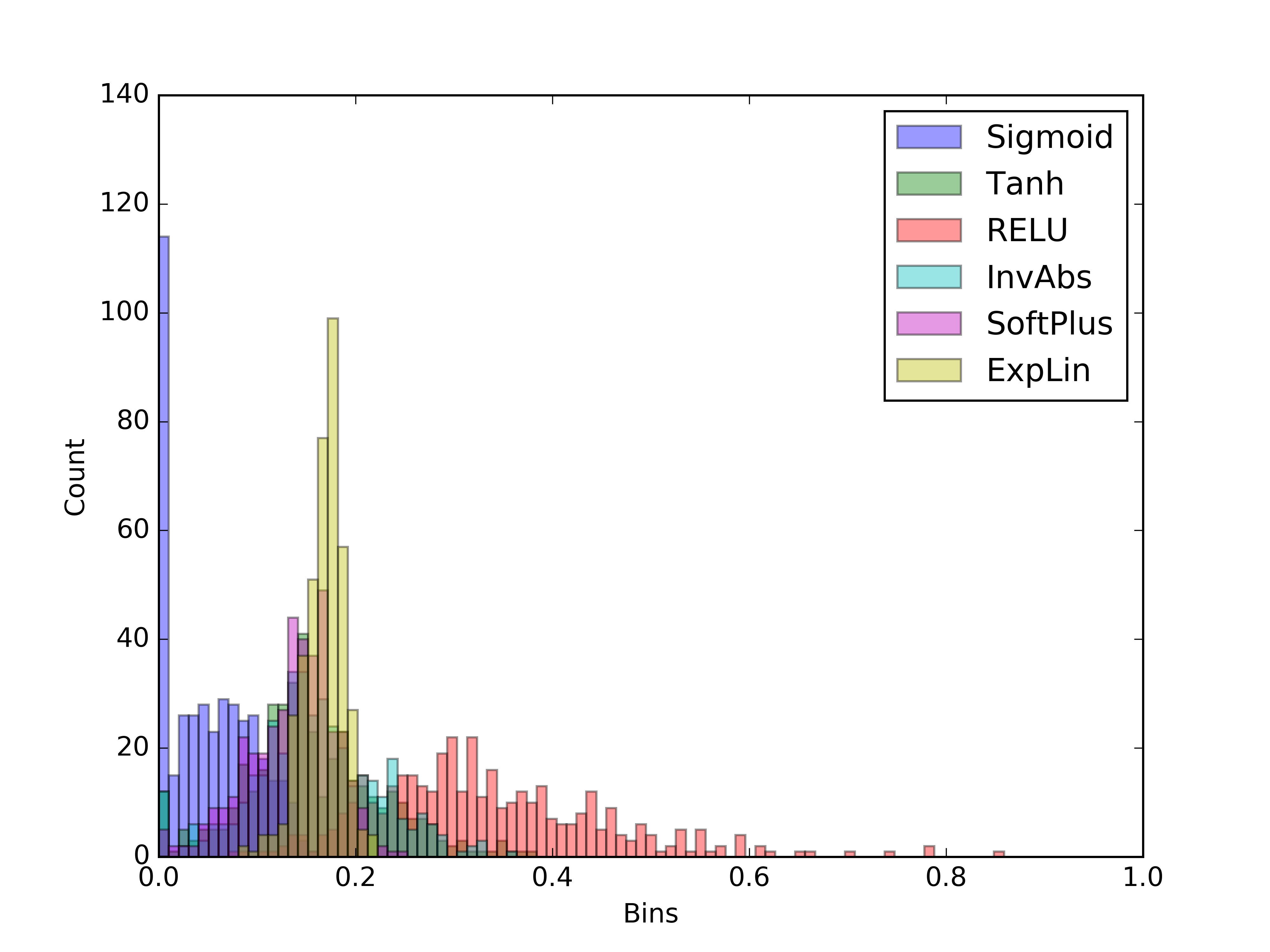}
  \end{minipage}
  \begin{minipage}[b]{0.22\textwidth}
    \includegraphics[width=\textwidth]{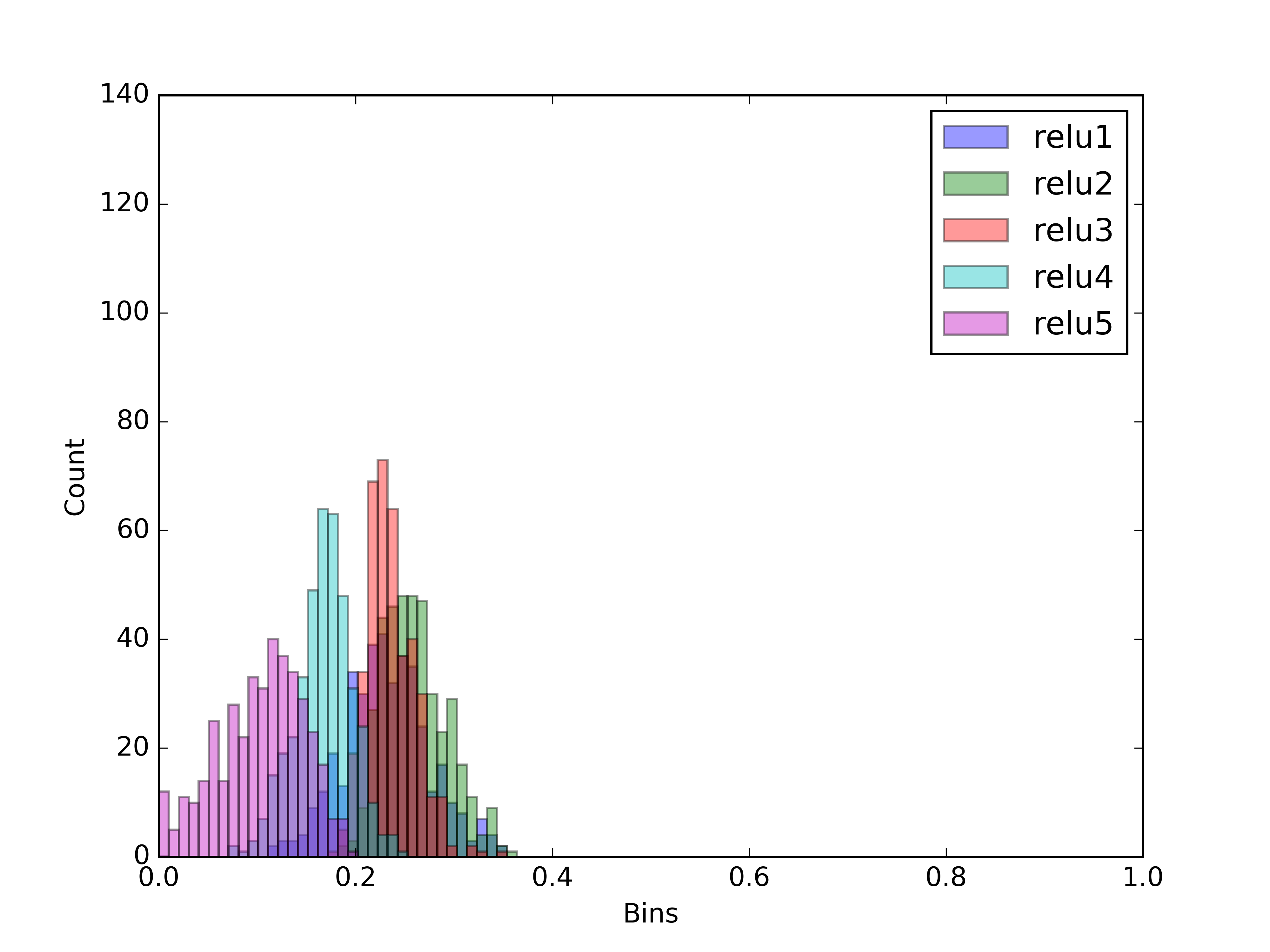}
  \end{minipage}
    \begin{minipage}[b]{0.22\textwidth}
    \includegraphics[width=\textwidth]{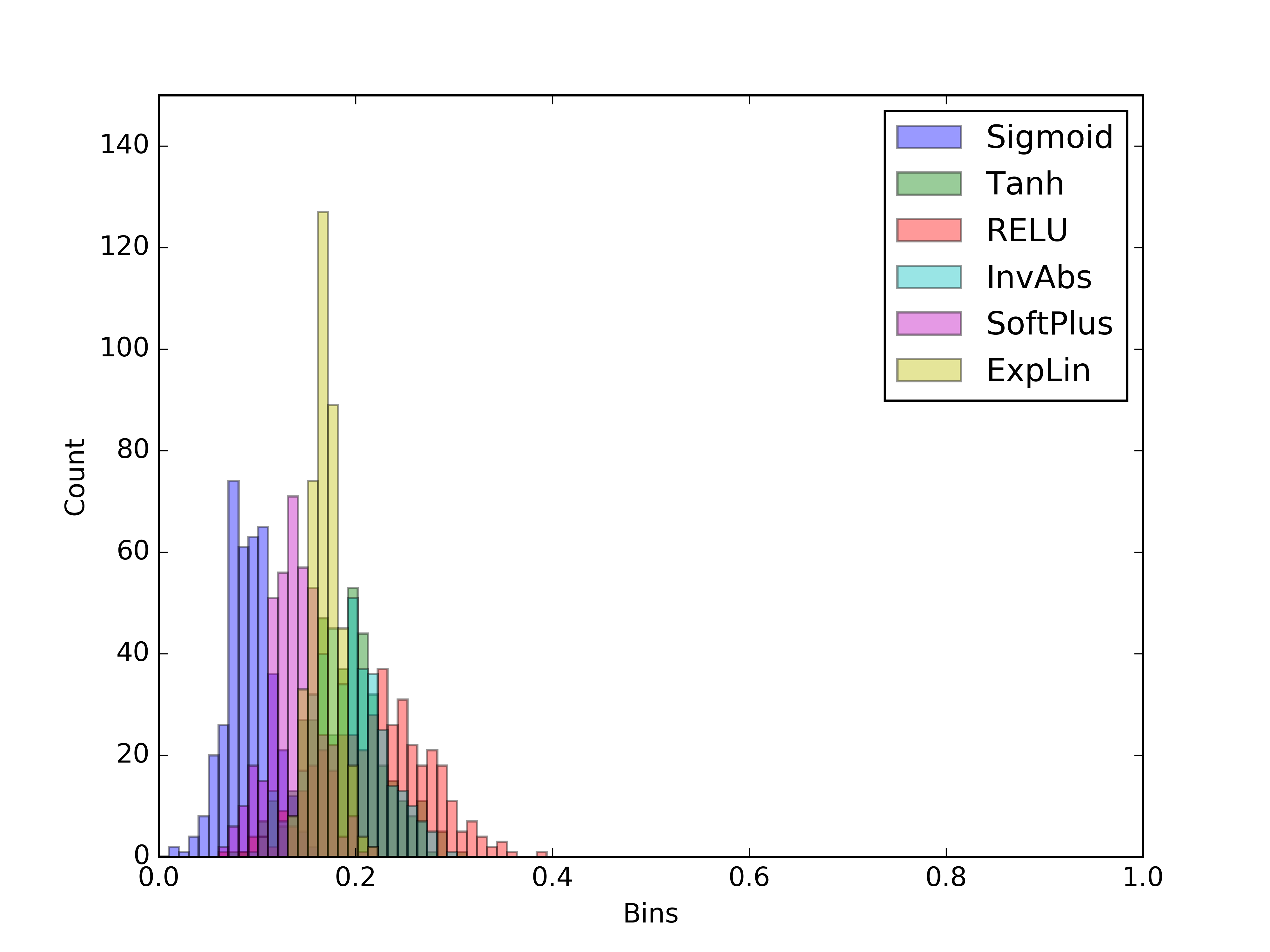}
  \end{minipage}
  \begin{minipage}[b]{0.22\textwidth}
    \includegraphics[width=\textwidth]{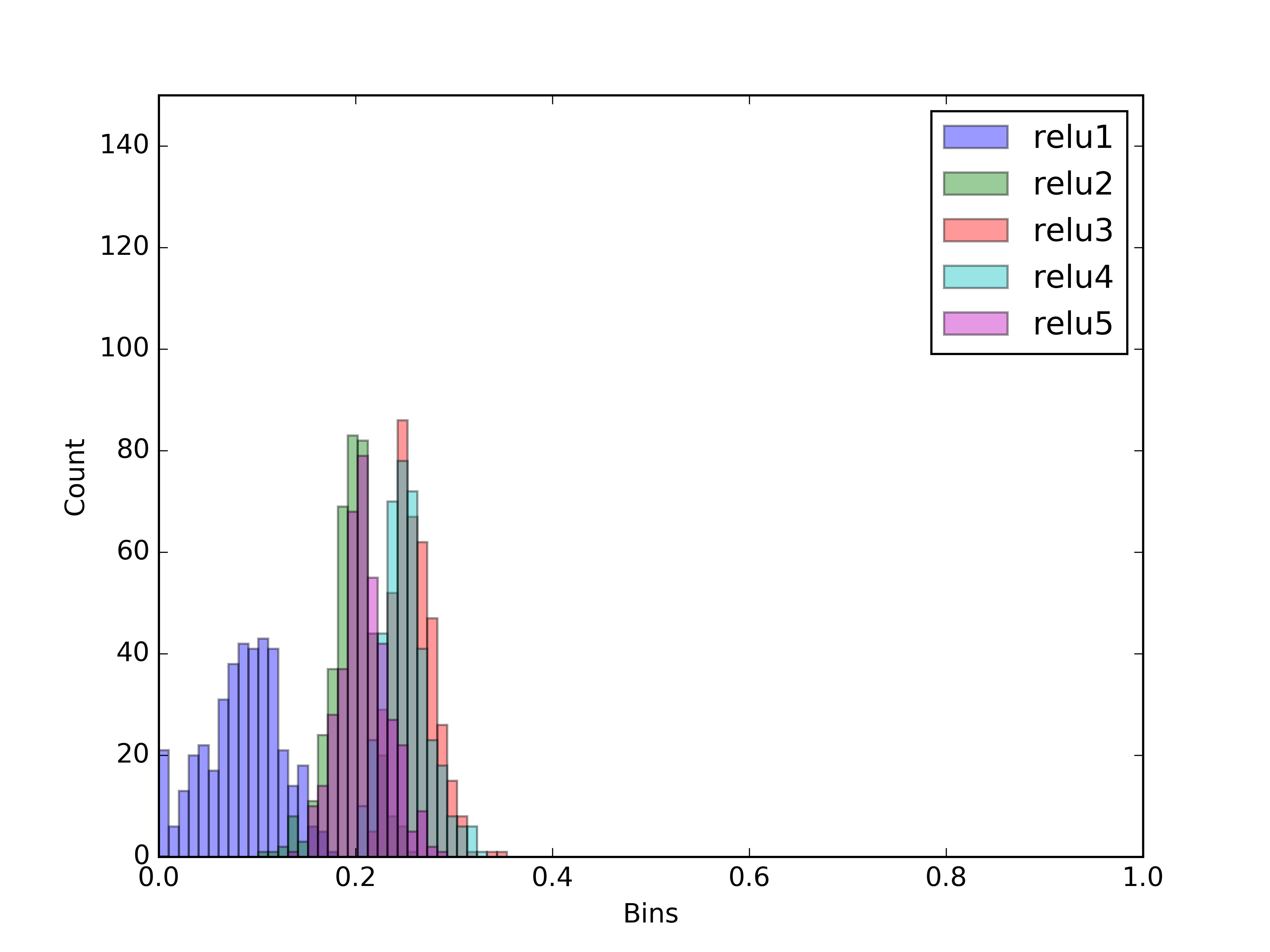}
  \end{minipage}
  \caption{Top to bottom layers 1-3 for a Feed Forward Network on MNIST.  Left is activation set 1 and right is activation set 2.}
\end{figure}

In Figure 3, we present histograms representing the share of $\alpha$ values for each activation function in the ensemble.  On the left are histograms representing the $\alpha$ values for each neuron when using the activations from set 1 and on the right are for the second set of activation functions.  The top left image in Figure 1 shows what we expect, namely that the ReLU dominates the layer.  As we move towards the bottom of the network, the ReLU is still the most favored function; however, the difference becomes marginal.  Also note that the sigmoid function is the lowest value function in all 3 layers and that the exponential linear function has the smallest range of all the functions.  On the right are the set of rectifier units with varying intercepts.  The dominant rectifier units in the first layer are the first and second rectifier units ($\max(0,x-1.0)$ and $\max(0,x-0.5)$) while the final layer is dominated by the third and fourth rectifier unit ($\max(0,x)$ and $\max(0,x+0.5)$).  Also observe that the most dominant activation set in our second set does not hold nearly as high values as seen in the first set.

\begin{figure}[h]
  \centering
  \begin{minipage}[b]{0.22\textwidth}
    \includegraphics[width=\textwidth]{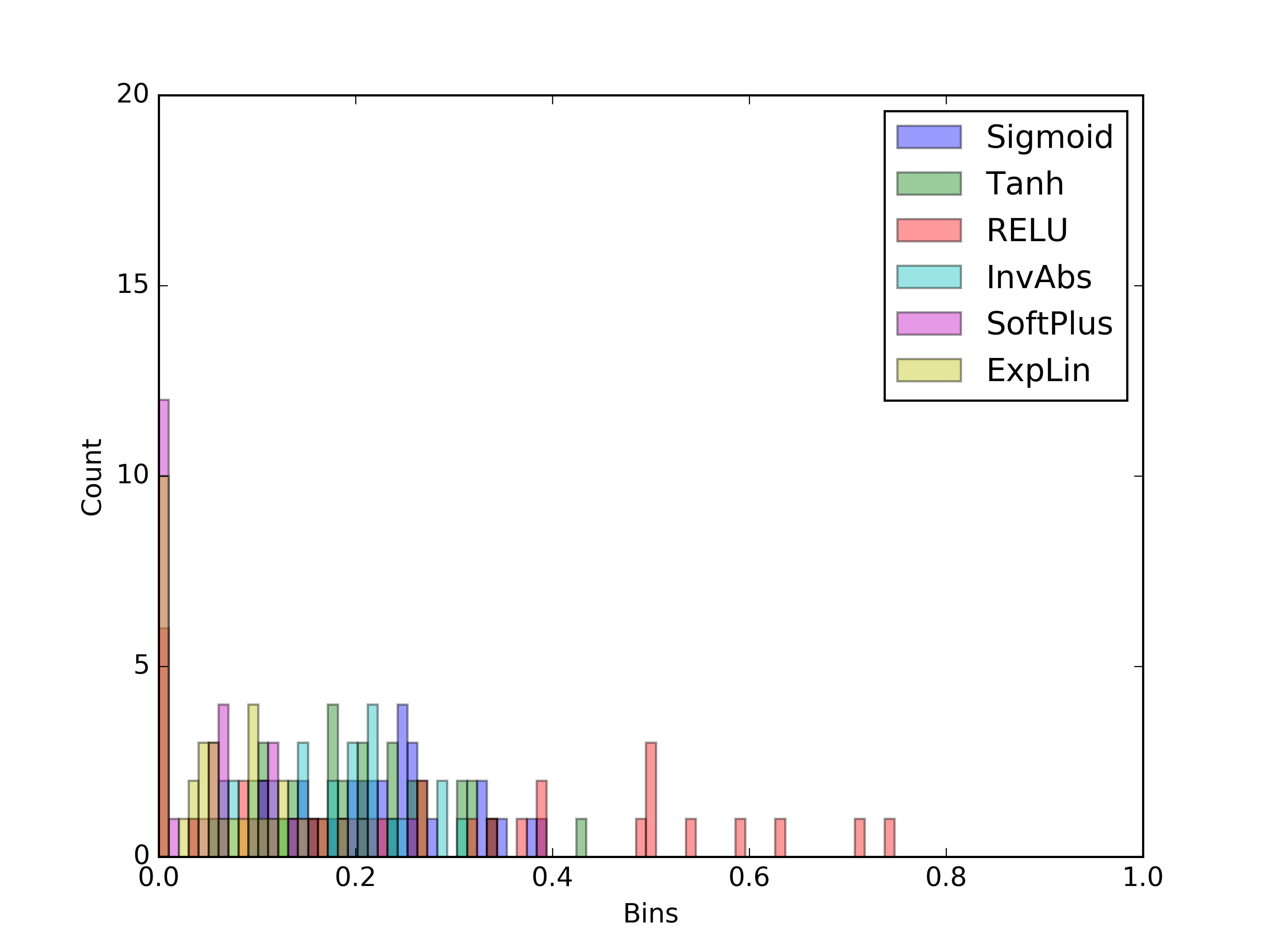}
  \end{minipage}
  \begin{minipage}[b]{0.22\textwidth}
    \includegraphics[width=\textwidth]{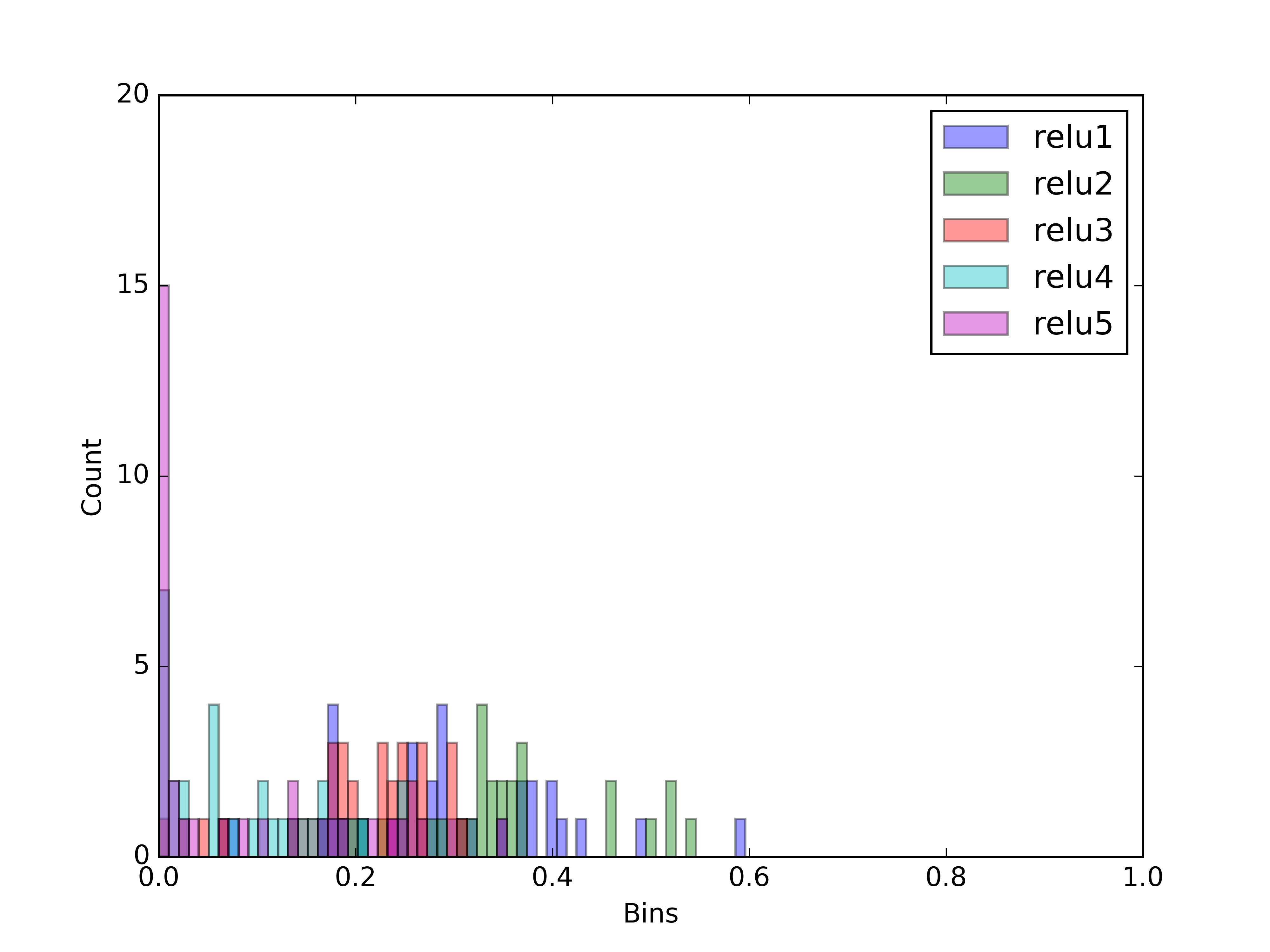}
  \end{minipage}
    \begin{minipage}[b]{0.22\textwidth}
    \includegraphics[width=\textwidth]{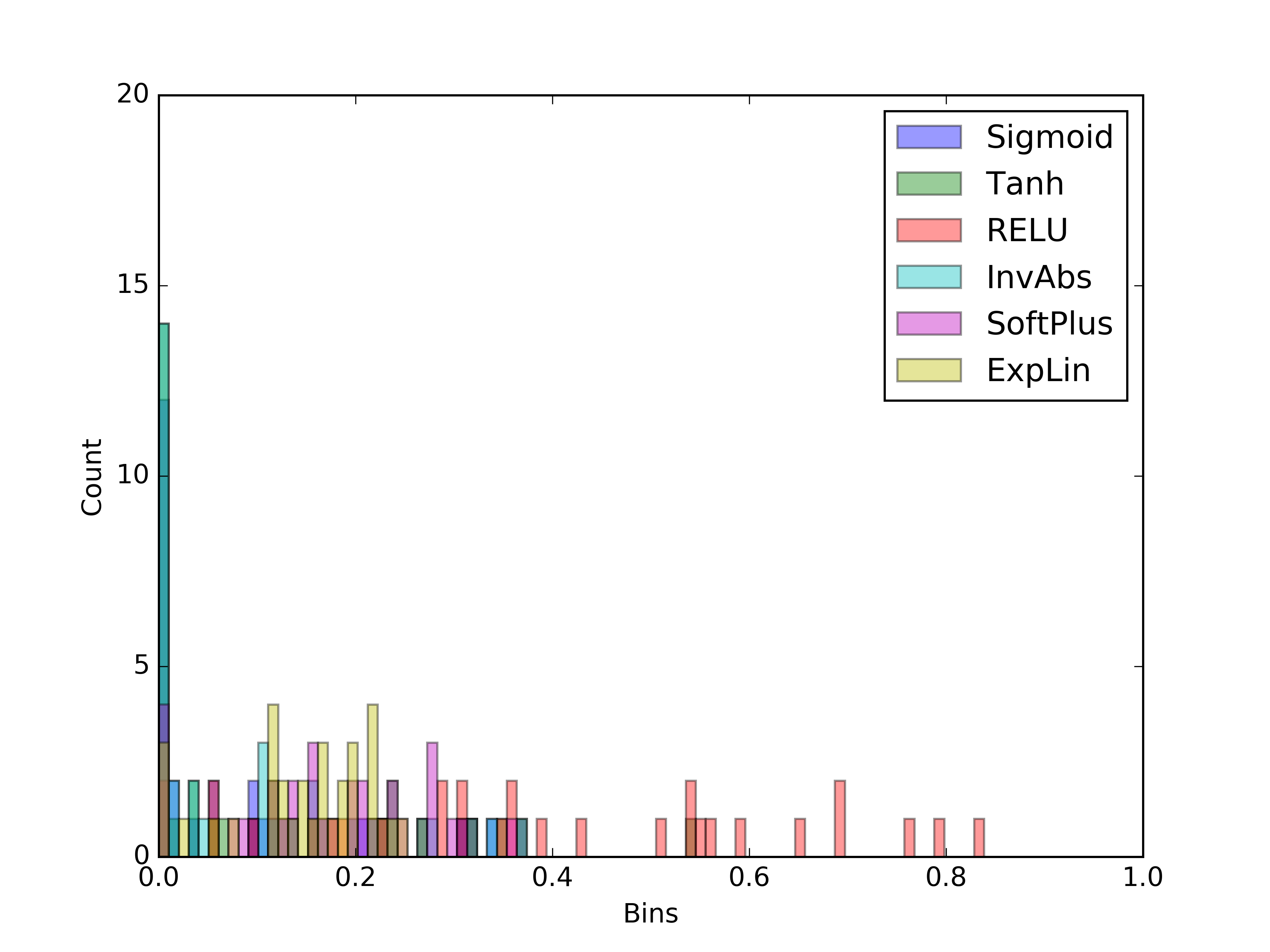}
  \end{minipage}
  \begin{minipage}[b]{0.22\textwidth}
    \includegraphics[width=\textwidth]{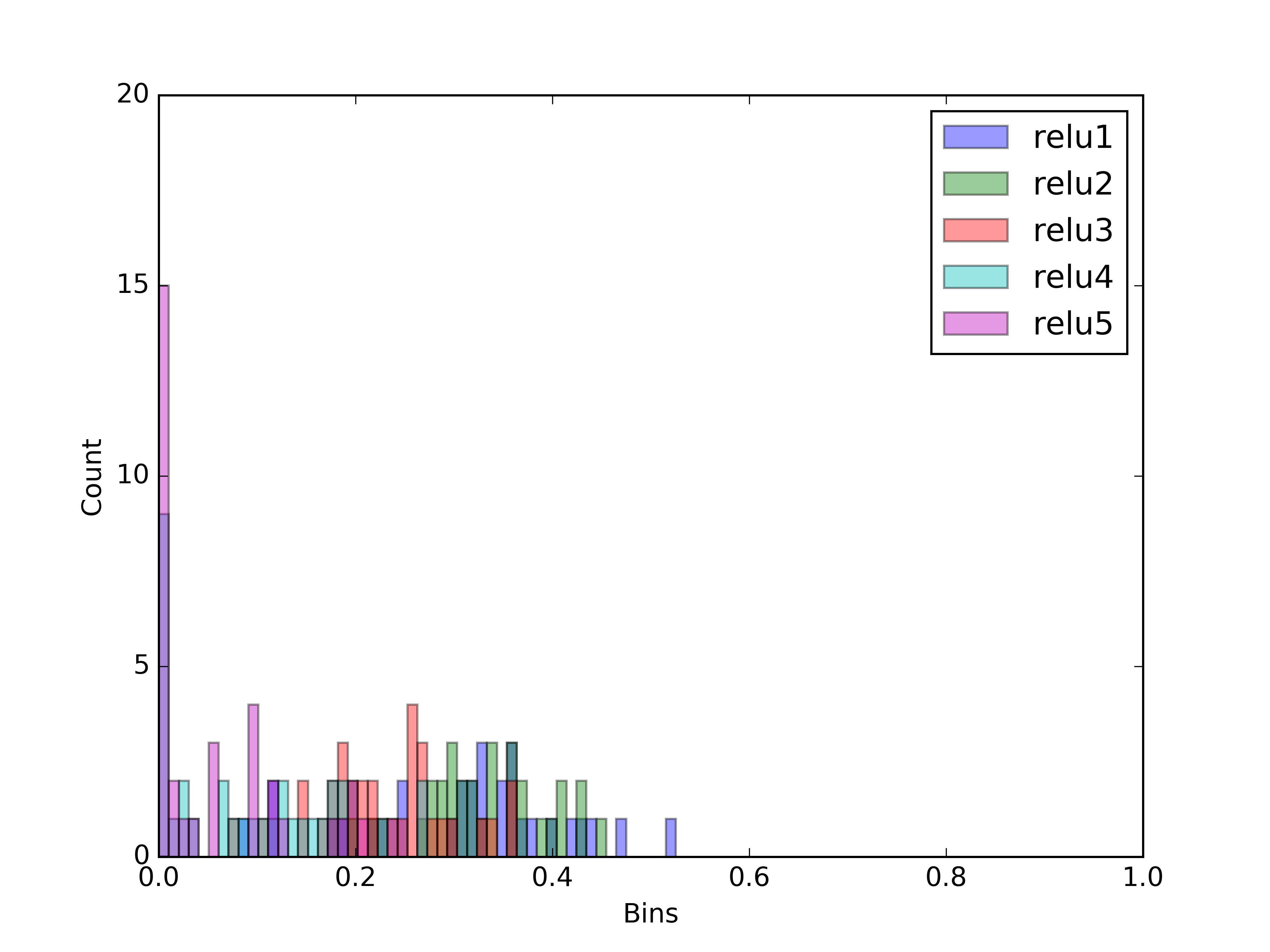}
  \end{minipage}
    \begin{minipage}[b]{0.22\textwidth}
    \includegraphics[width=\textwidth]{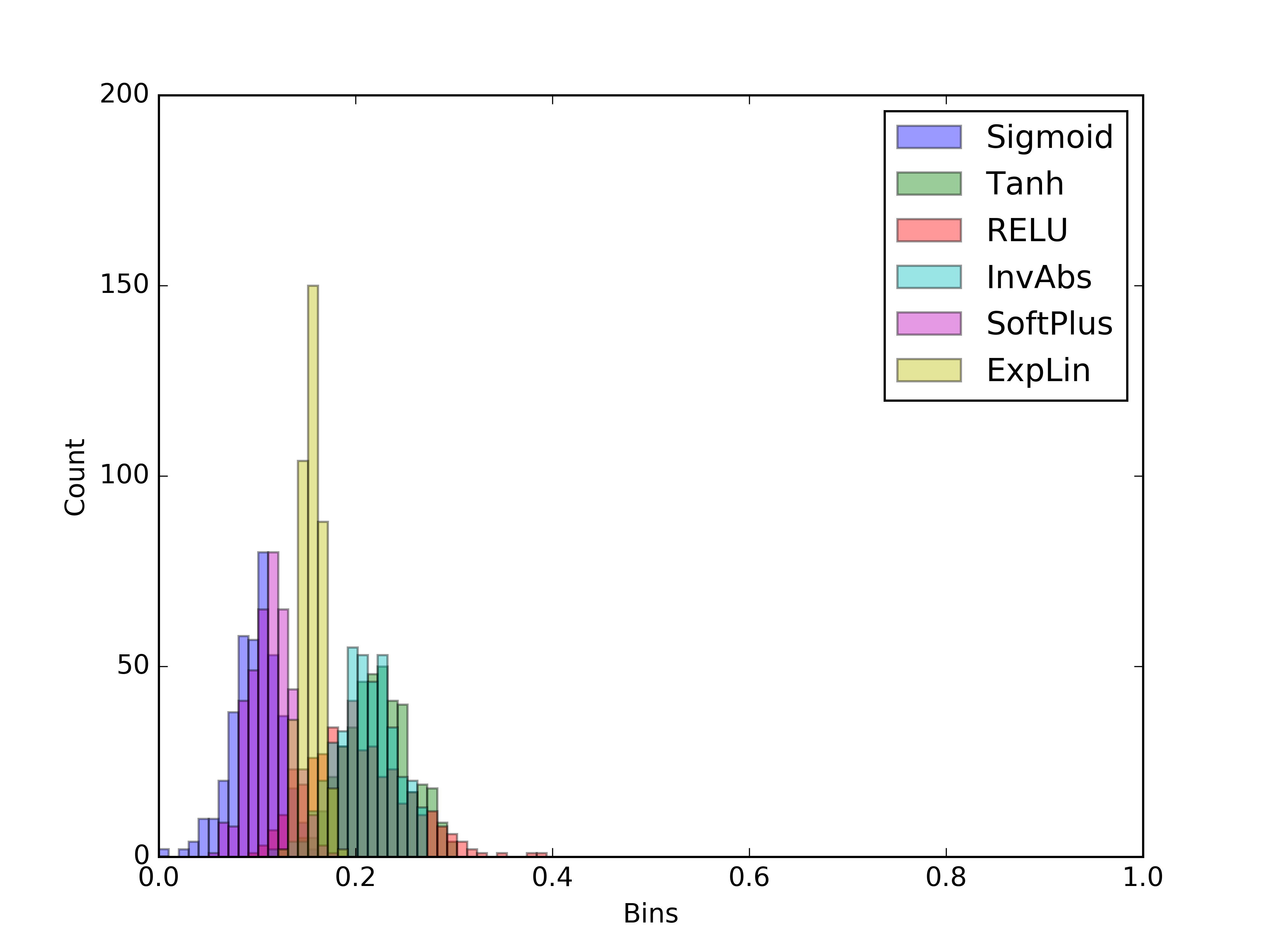}
  \end{minipage}
  \begin{minipage}[b]{0.22\textwidth}
    \includegraphics[width=\textwidth]{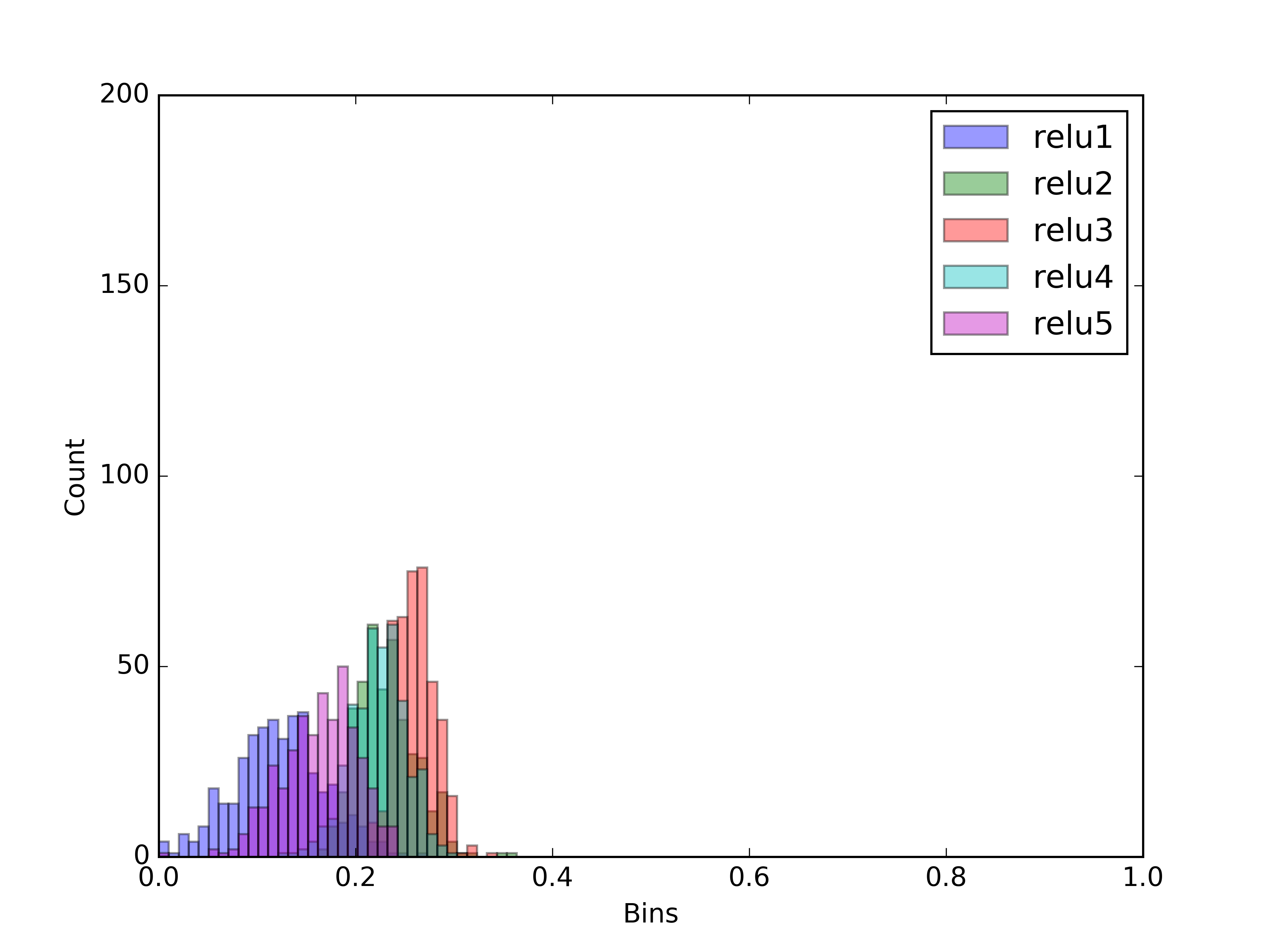}
  \end{minipage}
  \caption{Top to bottom layers 1-3 for a Convolutional Neural Network on MNIST.  Left is activation set 1 and right is activation set 2.}
\end{figure}

In Figure 4, we present the same histograms as Figure 3, but for the CNN model applied to MNIST.  Unsurprisingly, we observe that ReLU's dominate the top layer of the network. However, the story changes towards the bottom of the network. In fact, the hyperbolic tangent and inverse absolute value functions overtake the rectifier unit in importance. Interestingly, this third layer is also a feed forward layer after the prior convolutional layers. Thus, it appears that the transition of one type of layer to another changes the favored activation functions.  Another interesting behavior is for the second set of activations.  For the first two convolutional layers, there is not a dominant function. At the final activation ensemble layer, we find that the original ReLU is the chosen activation function.  Therefore, we see that given a single dataset, the favored activation depends on the layer and model used for classification.

\subsection{ISOLET}

ISOLET is a simple letter classification problem.  The data consists of audio data of people uttering letters.  The goal is to predict the letter said for each example (a-z).  ISOLET has 7797 examples and 617 attributes.  For training and testing, we split the data into 70\% for training and 30\% for testing.  In addition, we use our in-house FFN described in the previous section ($400f-400f-400f-26f$).  We train both a network with only ReLU's at each neuron and a network with activation ensembles.  For ISOLET, we implement all three classes of ensembles we mention in the previous section.

\begin{figure}[h]
  \centering
  \begin{minipage}[b]{0.22\textwidth}
    \includegraphics[width=\textwidth]{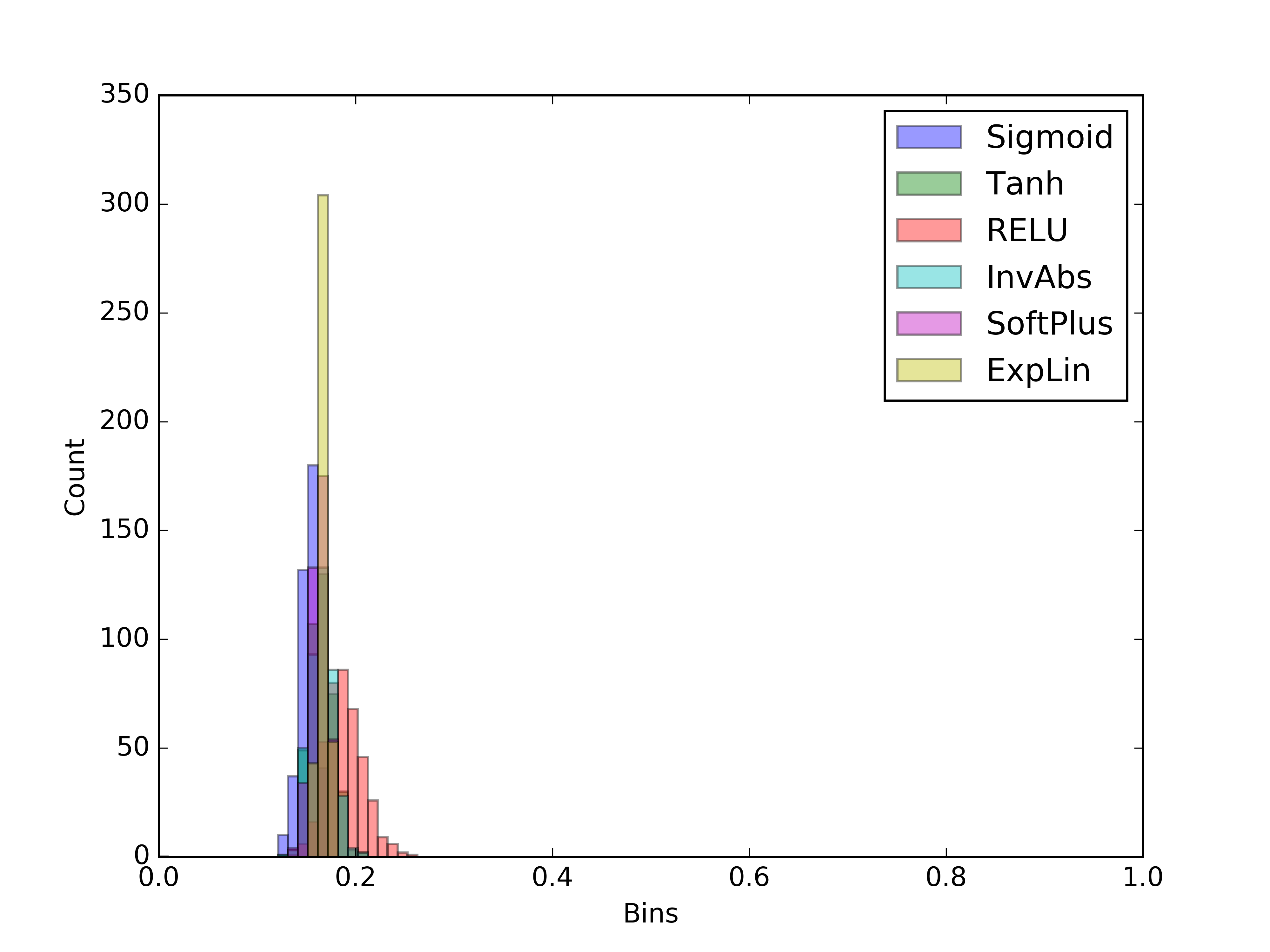}
  \end{minipage}
  \begin{minipage}[b]{0.22\textwidth}
    \includegraphics[width=\textwidth]{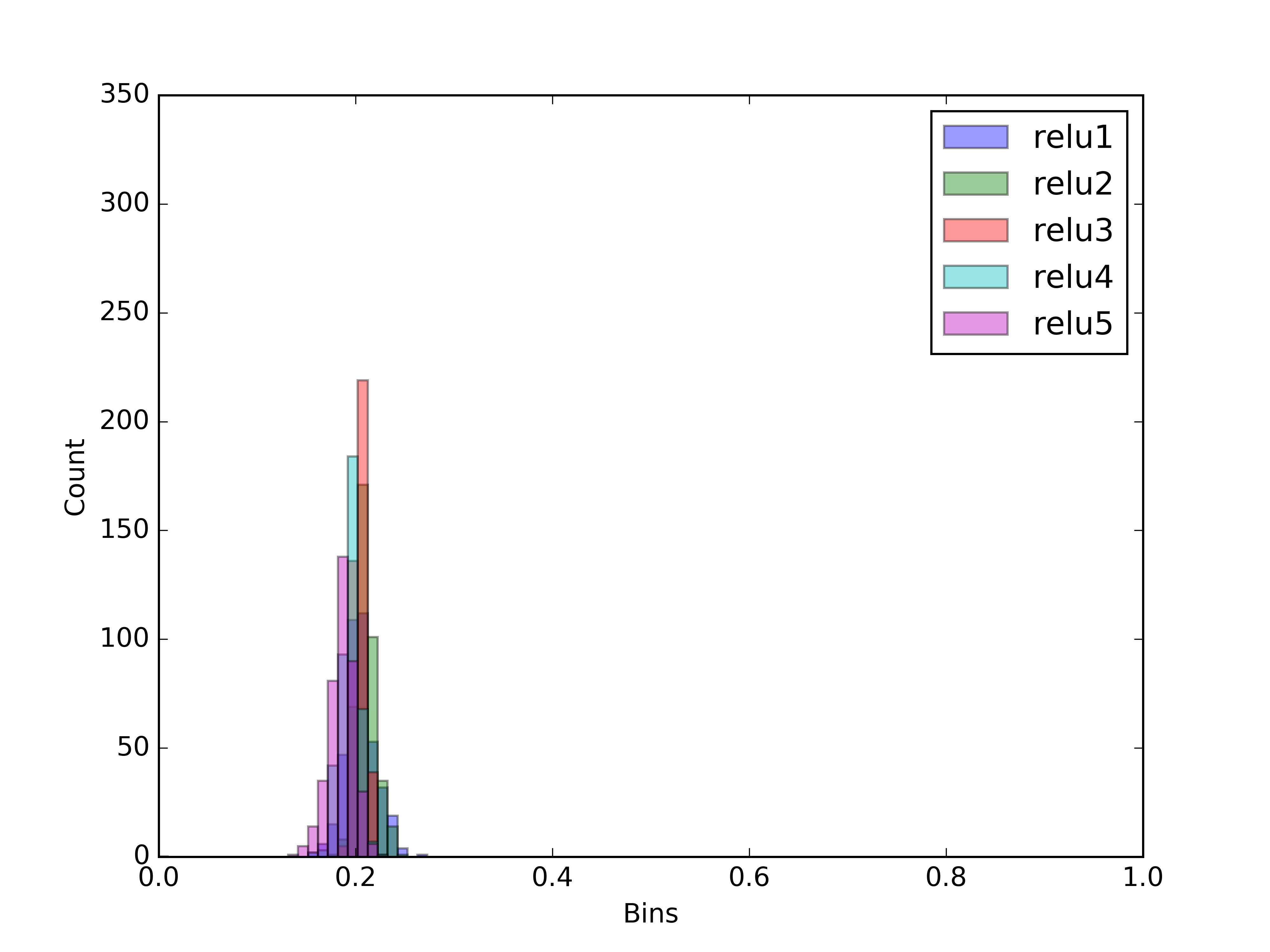}
  \end{minipage}
    \begin{minipage}[b]{0.22\textwidth}
    \includegraphics[width=\textwidth]{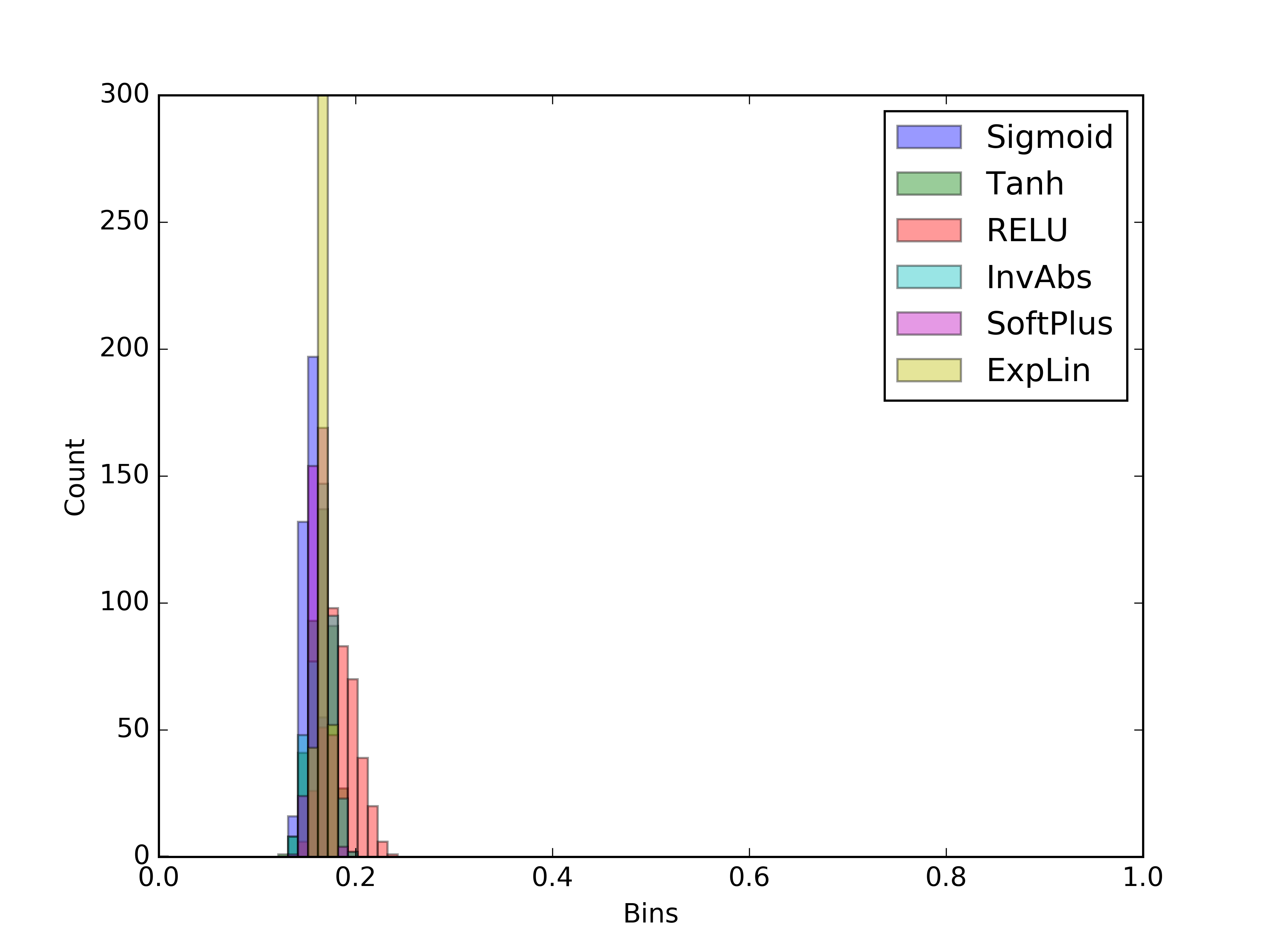}
  \end{minipage}
  \begin{minipage}[b]{0.22\textwidth}
    \includegraphics[width=\textwidth]{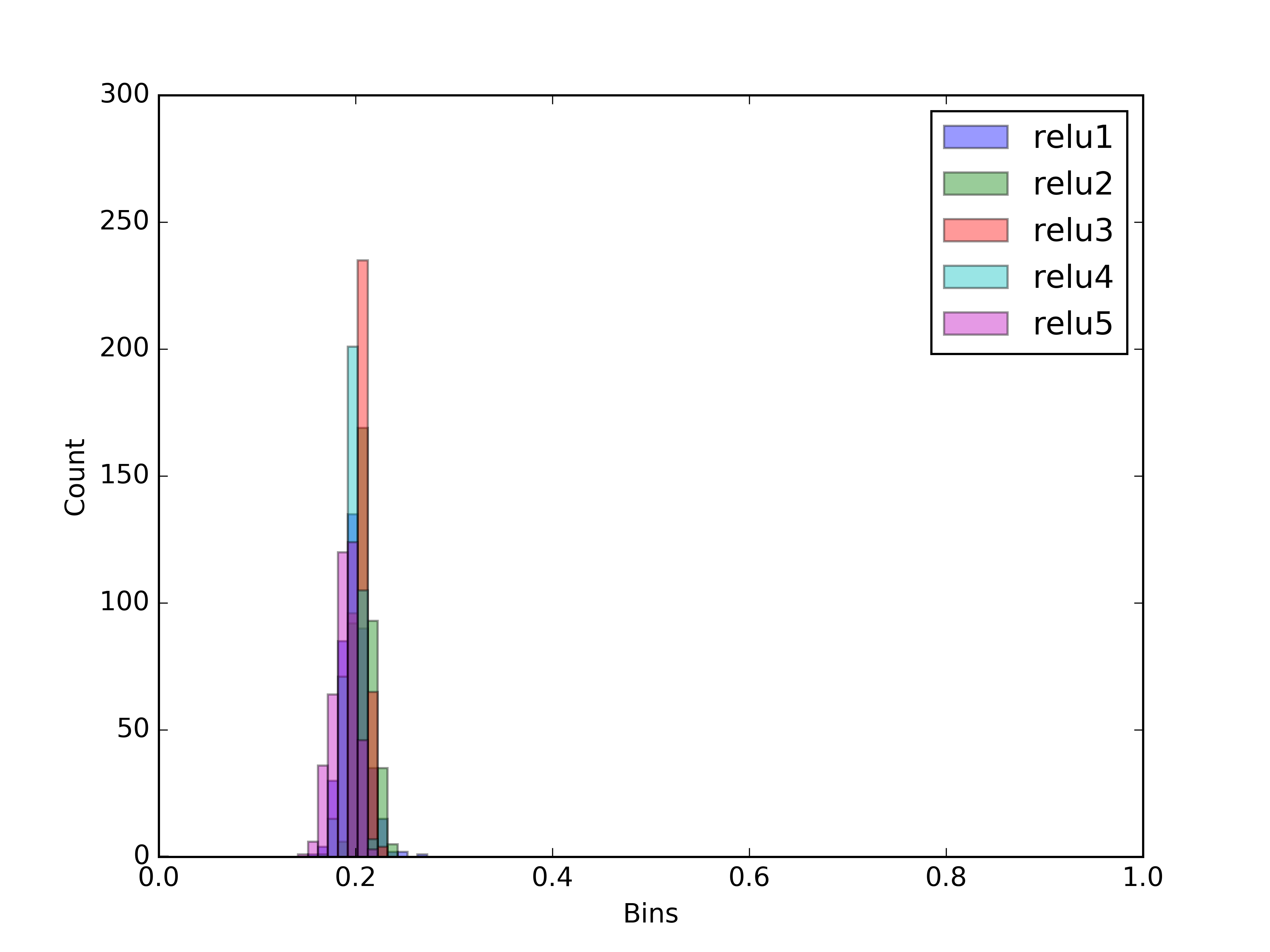}
  \end{minipage}
    \begin{minipage}[b]{0.22\textwidth}
    \includegraphics[width=\textwidth]{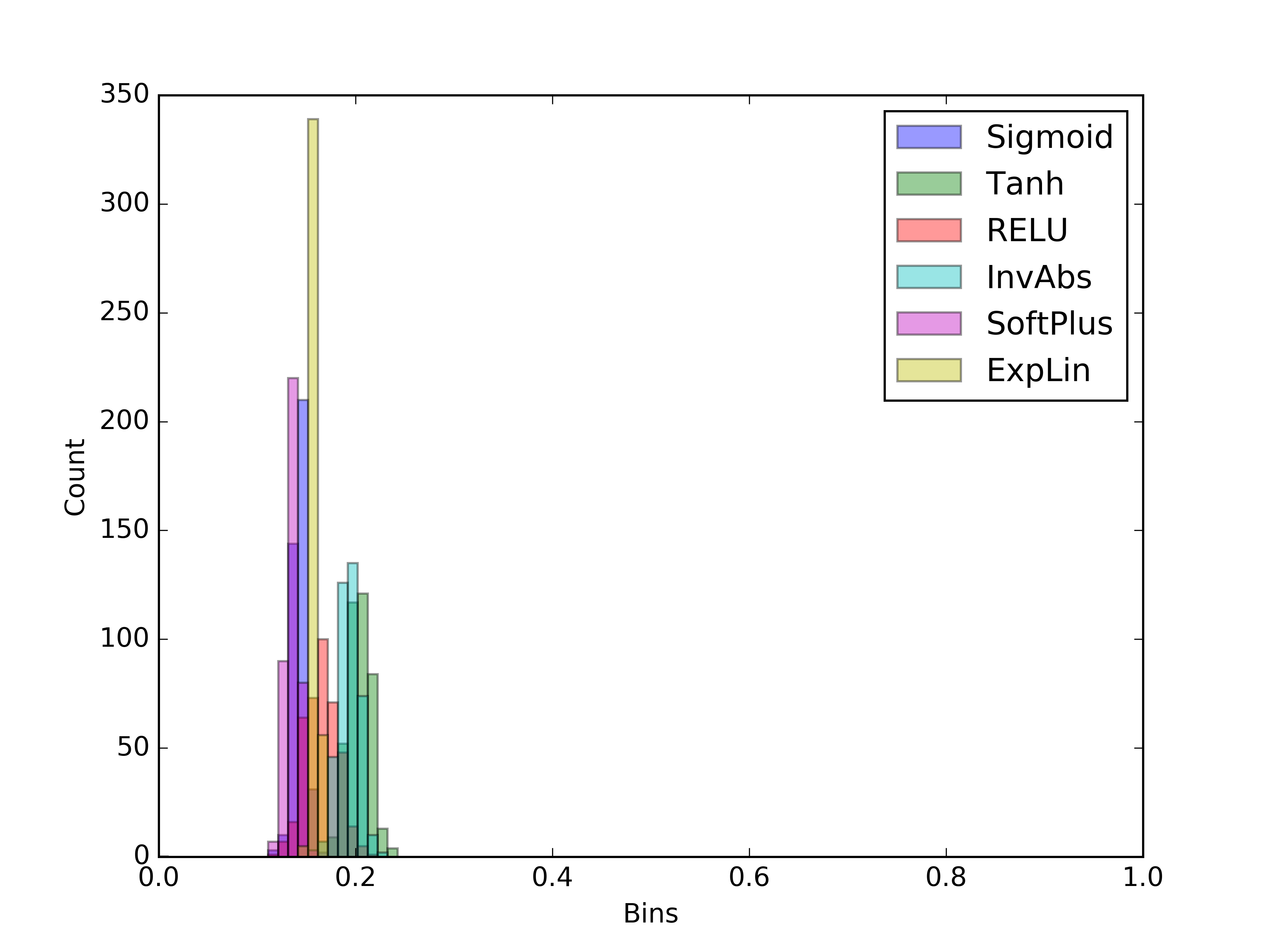}
  \end{minipage}
  \begin{minipage}[b]{0.22\textwidth}
    \includegraphics[width=\textwidth]{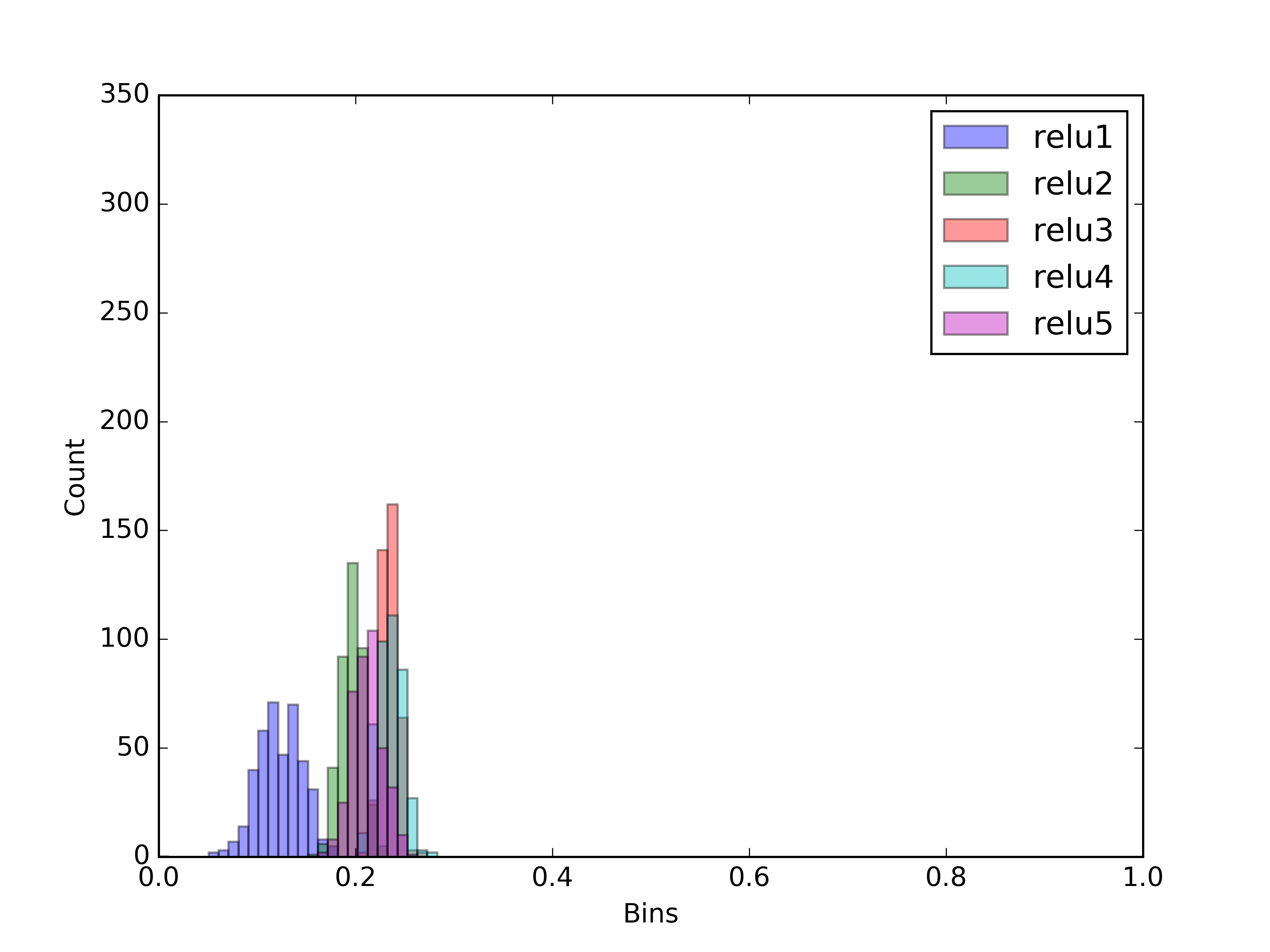}
  \end{minipage}
  \caption{Top to bottom layers 1-3 for a Feed Forward Network on ISOLET.  Left is activation set 1 and right is activation set 2.}
\end{figure}

In Figure 5, we present the same images provided for the MNIST dataset.  In the first set of images on the left, we see much different behavior of the $\alpha$ values than we did with the MNIST dataset.  Most values are clumped up near one another.  Though the ReLU is the leading activation function, it is not by much in the first two layers.  Similar to the CNN for MNIST, hyperbolic tangent and the inverse absolute value functions move up in importance.  For the second set of activation functions, we see that the $\alpha$ values are very close together and the network does not choose a dominant activation, even more so than for the first set.  However, at the final layer, the network singles out the first rectifier unit ($\max(0,x-1.0)$) as the worst of the set.  

Thus, we see that even with the same network and sets of activation functions, the model chooses different optimal activation functions based upon each individual dataset.

\subsection{CIFAR-100}

In order to get a broad scope of networks and test the strength of activation ensembles, we decide to use a residual network for CIFAR-100 since it both garners very good results for image-based datasets and offers and additional challenge for implementing the ensemble within its structure. The network we use comes from the Lasagne recipe Github \url{<https://github.com/Lasagne/Recipes/tree/master/papers>}.  It was originally designed for the CIFAR-10 dataset, but we modify the model for CIFAR-100.  The model has three residual columns of which include five residual block each followed by a global pooling layer. 

Challenges arise from two areas while implementing activation ensembles on residual networks.  First, residual networks are specifically designed for ReLU's via the residual element and initialization. Second, the placement of each activation ensemble needs to be carefully constructed to avoid disrupting the flow of information within each residual block.

We attempt various approaches to activation ensembles on residual networks.  The failed ideas include having an ensemble after each residual block and after the second stack but before adding in the residual. The implementation that works best is to incorporate the activation ensemble in the middle of a residual block and not at the end of the block or after.  In addition, this is the only network in which the second class of activations (the 5 ReLU's with various intercepts) works best. This is an expected result because of a residual network's dependence on the ReLU.

\subsection{STL-10}

The CAE we implement for this problem is very similar to one done by \citet{dosovitskiy2014discriminative}. However, our network contains less filters at each layer than their network due to insufficient GPU memory on our graphical cards.  Our structure is the following for the encoder portion of the autoencoder: $(32)5c-2p-(64)5c-2p-(128)5c-6p$.  Some CAE's use an ReLU at the end for prediction purposes, but we use the sigmoid function.  We keep normalization very simple with this network by only scaling the data by 255 at the input level.  The reproduced images by the network are then very easy for comparison by using the sigmoid function and mean-squared error.  During preliminary experiments we saw certain classes of activation functions tend to perform better, and thus we implemented only the third ensemble set that resembles an absolute value function.

We additionally tie all the weights for the encoder and decoder to restrain the model from simply finding the identity function.  Since tying the weights is suitable for autoencoders, we also tie the weights associated with our activation ensembles (this includes $\alpha, \delta,$ and $\eta$ values).  We choose to not tie the maximum and minimum values we find during the normalization stage as those are simply used for the purpose of our projection algorithm for the $\alpha$ variables.

\section{Conclusion}

In our work, we introduce a new concept we title as an activation ensemble.  Similar to common ensembling techniques in general machine learning, an activation ensemble is a combination of activation functions at each neuron in a neural network.  We describe the implementation for standard feed-forward networks, convolutional neural networks, residual networks, and convolutional autoencoders.  We create a convex combination of activation functions yet to be seen in literature with an algorithm to solve the new projection problem associated with our model.  In addition, we find that these ensembles improve classification accuracy for MNIST, ISOLET, CIFAR-100, and STL-10 (reconstruction loss).

To gain more insight into activation functions and their relationships with the neural network model implemented for a given dataset, we explore the $\alpha$'s for each activation function of each model.  Further, we examine one dataset, MNIST, with two models, FNN and CNN.  This enables us to discover that the optimal activation varies between a FFN and CNN on the same dataset (MNIST).  While the top two layers of both favor the rectifer unit, the bottom layer fancies the hyperbolic tangent and inverse absolute value functions.  When comparing the activation ensemble results between ISOLET and MNIST on the FFN, we observe that the optimal activation functions between datasets are also different.  While MNIST esteems the ReLU, ISOLET does not favor a particular activation until the bottom layer, in which it chooses the hyperbolic tangent.

\bibliographystyle{icml2017}
\bibliography{zpaper}
\nocite{lichman2013uci}
\nocite{krizhevsky2009learning}
\nocite{coates2010analysis}
\nocite{lecun2015deep}

\end{document}